\documentclass[journal]{IEEEtran}
\ifCLASSOPTIONcompsoc
  \usepackage[nocompress]{cite}
\else
  \usepackage{cite}
\fi
\usepackage{amsmath,amssymb,bm,mathrsfs}
\usepackage{booktabs,multirow,makecell}
\usepackage{graphicx}
\usepackage{algorithm}
\usepackage{algpseudocode}
\usepackage{enumitem}
\usepackage[table]{xcolor}
\usepackage{array}
\usepackage{url}

\usepackage{float}
\usepackage{placeins}
\usepackage{dblfloatfix}
\usepackage{microtype}
\usepackage{comment}
\usepackage{hyperref}

\graphicspath{{./}{../}}
\DeclareGraphicsExtensions{.pdf,.png,.jpg,.jpeg}

\newcommand{\ourmethod}{\textsc{UCGP}}
\newcommand{\best}[1]{\textbf{#1}}
\newcommand{\second}[1]{\underline{#1}}
\newcommand{\runinhead}[1]{\textbf{#1}\nobreak\hspace{0.35em}\ignorespaces}
\definecolor{ACMBurgundy}{HTML}{8F1D2C}
\definecolor{OursGray}{HTML}{ECECEC}
\definecolor{DiagGray}{HTML}{D9D9D9}
\definecolor{CorrectGreen}{HTML}{538234}
\definecolor{ErrorRed}{HTML}{D6171C}
\definecolor{SoftOrange}{HTML}{E6A23C}
\definecolor{TPAMILinkBlue}{HTML}{1F4E79}
\definecolor{TPAMIUrlBlue}{HTML}{1B6A7A}
\hypersetup{
  colorlinks=true,
  linkcolor=TPAMILinkBlue,
  citecolor=TPAMILinkBlue,
  urlcolor=TPAMIUrlBlue,
  pdfauthor={Chengyin Hu, Yuxian Dong, Yikun Guo, Xiang Chen, Qike Zhang, Junqi Wu, Jiahuan Long, Jiujiang Guo, Yiwei Wei, Tingsong Jiang, and Wen Yao},
  pdftitle={Revealing Physical-World Semantic Vulnerabilities: Universal Adversarial Patch for Infrared Vision-Language Models}
}
\newcommand{\dec}[1]{\textcolor{ACMBurgundy}{\,\(\downarrow\)#1}}
\newcommand{\correct}[1]{\textcolor{CorrectGreen}{#1}}
\newcommand{\wrong}[1]{\textcolor{ErrorRed}{#1}}

\newcommand{\diagcell}[1]{\cellcolor{DiagGray}\best{#1}}
\newcommand{\oursheader}{\textbf{\ourmethod{} (Ours)}}
\newcommand{\srcdomain}[1]{\ensuremath{\mathcal{D}_{#1}}}
\newcommand{\Lsub}{\mathcal{L}_{\text{subspace}}}
\newcommand{\Ltop}{\mathcal{L}_{\text{topo}}}
\newcommand{\Lbud}{\mathcal{L}_{\text{budget}}}
\algrenewcommand\algorithmicrequire{\textbf{Input:}}
\algrenewcommand\algorithmicensure{\textbf{Output:}}
\newcolumntype{L}[1]{>{\centering\arraybackslash}m{#1}}
\newcolumntype{M}[1]{>{\centering\arraybackslash}m{#1}}
\newcolumntype{C}[1]{>{\centering\arraybackslash}m{#1}}
\newcolumntype{G}[1]{>{\columncolor{OursGray}[\tabcolsep][\tabcolsep]\centering\arraybackslash}m{#1}}

\begin{document}
\bstctlcite{IEEEexample:BSTcontrol}

\title{Revealing Physical-World Semantic Vulnerabilities: Universal Adversarial Patch for Infrared Vision-Language Models}

\author{Chengyin Hu, Yuxian Dong, Yikun Guo, Xiang Chen, Qike Zhang, Junqi Wu, Jiahuan Long\textsuperscript{\textdagger},
Jiujiang Guo, Yiwei Wei\textsuperscript{\textdagger}, Tingsong Jiang, and Wen Yao%
\thanks{Chengyin Hu, Yuxian Dong, Yikun Guo, Xiang Chen, Qike Zhang, and Yiwei Wei are with China University of Petroleum-Beijing at Karamay, No. 355 Anding Road, Karamay 834000, Xinjiang, China. (e-mail: cyhu@cupk.edu.cn; 854531750@st.cupk.edu.cn; gyk666@st.cupk.edu.cn; cxcx@st.cupk.edu.cn; zhangqike@st.cupk.edu.cn; weiyiwei@cupk.edu.cn).}%
\thanks{Junqi Wu, Tingsong Jiang, and Wen Yao are with the Defense Innovation Institute, Chinese Academy of Military Science, Beijing, China. (e-mail: wujq28@sjtu.edu.cn; tingsong@pku.edu.cn; wendy0782@126.com).}%
\thanks{Jiahuan Long is with Shenzhen University, Shenzhen, China. (e-mail: jiahuanlong@sjtu.edu.cn).}%
\thanks{Jiujiang Guo is with North University of China, Taiyuan, China (e-mail: gjiujiang@163.com).}%
\thanks{\textsuperscript{\textdagger}Corresponding authors: Jiahuan Long and Yiwei Wei.}}

\markboth{MANUSCRIPT FOR IEEE TRANSACTIONS ON PATTERN ANALYSIS AND MACHINE INTELLIGENCE}%
{Hu \MakeLowercase{\textit{et al.}}: Revealing Physical-World Semantic Vulnerabilities}

\maketitle

\begin{abstract}
Infrared vision-language models (IR-VLMs) are becoming important for semantic perception in low-visibility environments, yet their robustness to physical semantic attacks remains underexplored. Existing adversarial patch methods are mainly designed for red-green-blue (RGB) images or closed-set infrared detectors and do not directly address open-ended IR-VLM tasks, where one deployable artifact can affect classification, captioning, and visual question answering (VQA) simultaneously. We propose Universal Curved-Grid Patch, abbreviated \ourmethod{}, a universal physical adversarial patch framework tailored to IR-VLMs. \ourmethod{} represents the patch as a deployable low-frequency curved grid and optimizes it with a representation-driven objective over subspace departure, topology disruption, and local appearance regularization. Meta Differential Evolution (MetaDE) searches for patch parameters with two physical robustness augmentations: Expectation over Transformation (EOT) capture sampling for imaging variations and thin-plate spline (TPS) patch-deformation modeling for non-rigid local shape changes. Rather than manipulating labels or prompts, \ourmethod{} disrupts the clean-category manifold in visual representation space, which later appears as degraded cross-modal outputs. Experiments show that a single shared patch degrades classification, captioning, and VQA across diverse IR-VLM architectures while retaining measurable cross-model transfer, cross-dataset generalization, cross-category extensibility, and real-scene physical effectiveness under the evaluated conditions. These results reveal a robustness blind spot in current infrared multimodal systems. Code and reproducibility assets are available at \url{https://github.com/dyx6663/UCGP}.
\end{abstract}

\begin{IEEEkeywords}
Infrared vision-language models, Physical adversarial attack, Universal curved-grid patch
\end{IEEEkeywords}
\section{Introduction}
\label{sec:introduction}

\IEEEPARstart{V}{ision-language} models, or VLMs, achieve strong performance on retrieval, captioning, and visual question answering by aligning visual and textual representations\cite{radford2021clip}. As this paradigm extends to infrared perception, recent work has introduced datasets, models, and benchmarks for infrared semantic understanding\cite{jiang2024infraredllava,cao2025irgpt,moshtaghi2025rgbthbench,zhang2025ifbench}. In this paper, IR-VLM denotes a VLM backbone adapted to infrared tasks such as classification, captioning, and VQA. This shift enlarges the attack surface from recognition errors to semantic deviations in generated descriptions and answers.

\begin{figure}[t]
\centering
\includegraphics[width=0.88\columnwidth]{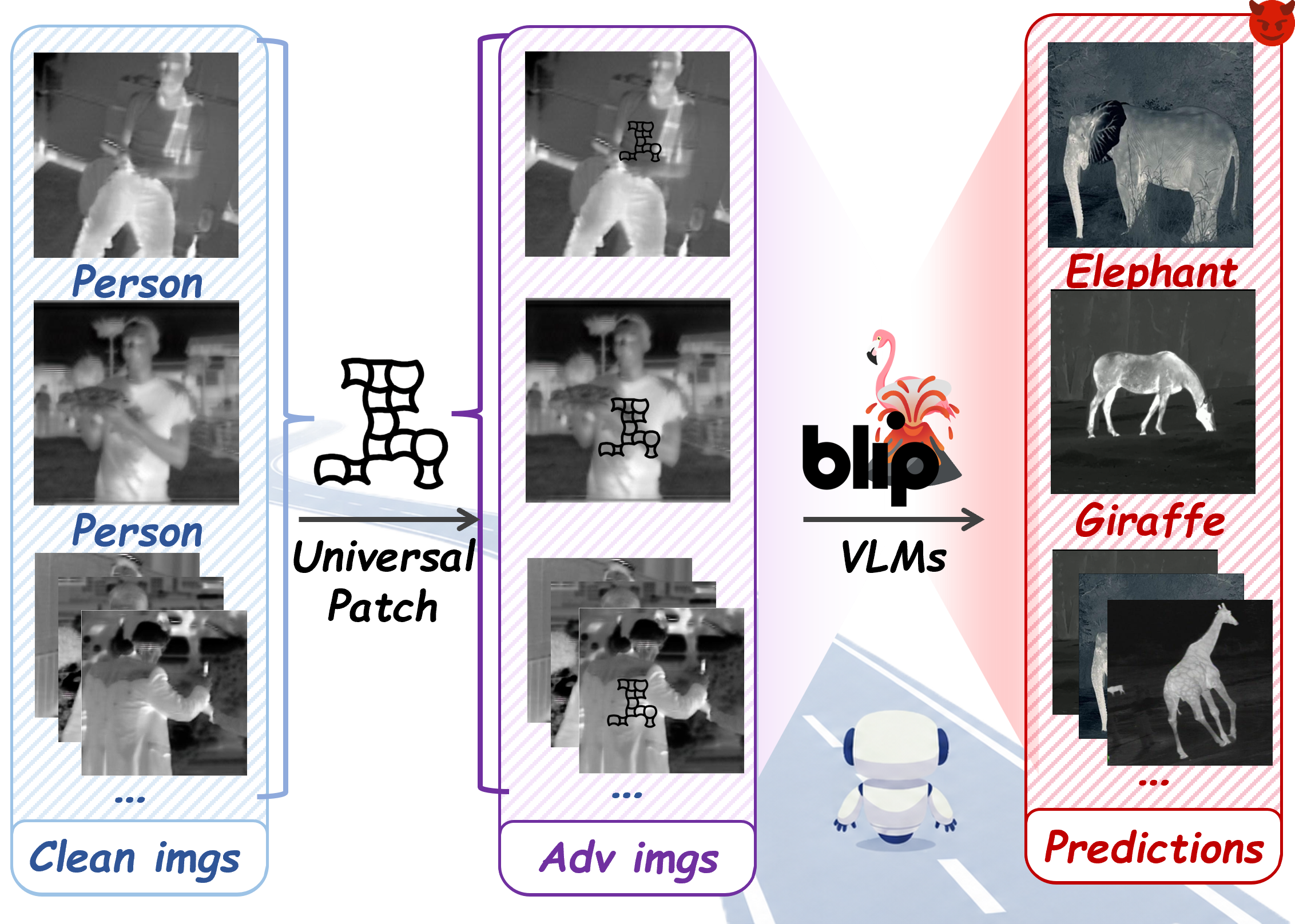}
\caption{Universal-patch attack overview. A single shared curved-grid patch is attached to the target region of different clean infrared images and reused without per-image re-optimization. The resulting adversarial inputs alter the visual evidence provided to IR-VLMs.}
\label{fig:attack_overview}
\end{figure}
 
Prior work has shown that carefully designed perturbations can deceive deep visual models in both digital and physical settings\cite{athalye2018eot,eykholt2018rp2}. As multimodal perception evolves from closed-set recognition to open-ended semantic understanding, the security question also changes: failures no longer stop at wrong labels, but can propagate into generated descriptions and answers. For infrared vision-language systems, this risk remains largely uncharacterized. Existing efforts provide only a partial picture, and it is still unclear whether one physically deployable artifact can manipulate the shared visual evidence used by classification, captioning, and VQA once infrared perception is coupled with language reasoning\cite{qi2024visualjailbreak,xie2025chainattack,wei2023unified,hu2024infraredcurves,tiliwalidi2024infraredgrid}. This gap motivates the central question of this paper: can a universal physical patch induce stable open-ended semantic deviations in IR-VLMs? We answer this question by formulating and validating a universal physical adversarial patch setting for IR-VLMs.

The universal-patch setting is motivated by the deployment constraint rather than by convenience alone. A physical attack cannot normally fabricate a different patch for every frame, prompt, or downstream task; it must attach one object to a target region and reuse it under viewpoint, distance, and background changes. At the same time, infrared images exhibit weaker texture and noise-dominated statistics\cite{hou2022infraredreview,jia2021llvip}, so visible-domain high-frequency patterns may either disappear after thermal capture or become difficult to fabricate. These constraints suggest that an effective IR-VLM attack should be shared, low-frequency, region-constrained, and tied to the visual representation before task-specific decoding. To address these challenges, we propose \ourmethod{}, a universal physical adversarial patch framework tailored to IR-VLMs. \ourmethod{} integrates deployment-friendly patch parameterization, representation-level objective design, and physical robustness modeling. It learns a single shared patch that can be directly reused across samples without per-sample re-optimization. Fig.~\ref{fig:attack_overview} summarizes the shared-patch deployment pipeline and the resulting semantic deviations across classification, captioning, and VQA.

We validate \ourmethod{} through classification, captioning, VQA, transfer, physical, defense, and cross-category experiments, with three main contributions:
\begin{itemize}[leftmargin=1.5em,topsep=4pt,itemsep=2pt]
\item To the best of our knowledge, we are the first to propose and systematically study a universal physical adversarial patch framework tailored to IR-VLMs, establishing a unified setting built around a shared deployable patch, cross-sample regional attachment, and physical robustness.
\item We conduct extensive experiments on classification, captioning, VQA, transfer, defense, and physical deployment. Under the evaluated protocols, UCGP achieves strong attack effectiveness across digital and physical settings and consistently outperforms the infrared physical baselines considered in this study.
\item Through analyses on captioning, VQA, transfer, defense, cross-category extension, and physical deployment, we show that a single shared patch can induce semantic degradation beyond closed-set label flipping, while also exposing limitations under defenses and real-scene capture conditions.
\end{itemize}

\section{Related Work}
\label{sec:related-work}

\runinhead{Infrared physical attacks.} Physical adversarial patches and robust physical attacks reveal that deployable perturbations can remain effective after capture, viewpoint change, and scale variation\cite{goodfellow2015fgsm,brown2017adversarial,athalye2018eot,eykholt2018rp2,chen2018shapeshifter}. Compared with digital perturbations, physical patches must survive material fabrication, local deformation, camera noise, scale changes, thermal drift, and imperfect placement. These constraints are sharper in infrared imaging because sensors record thermal radiation rather than visible texture: an RGB-clear pattern may have weak thermal contrast, whereas a thermally strong patch may remain visually unobtrusive~\cite{hou2022infraredreview}.

Prior infrared physical attacks have been developed mainly for pedestrian detectors. Early studies used adversarial clothing or insulating materials to hide thermal signatures\cite{zhu2022invisibleclothing,wei2023infraredpatch}; later works explored attachable structures such as blocks, curves, and grids to improve deployability and multi-view robustness\cite{hu2023infraredblocks,hu2024infraredcurves,tiliwalidi2024infraredgrid}. Most closely, Wei et al.\cite{wei2023unified} proposed one shape-optimized physical patch that simultaneously attacks visible--infrared detector pairs, using score-aware optimization to balance evasion across the two sensor modalities. This is cross-modal at the sensor level, whereas our setting is cross-modal at the vision--language level: we perturb an infrared visual representation and evaluate the resulting semantic errors in classification, captioning, and VQA. Thus, these works provide important deployment priors, but their primary target remains closed-set detection or recognition. Our setting differs in both objective and evaluation: one reusable patch should disturb a shared visual representation before downstream language generation, and the resulting failure should be tested across classification, captioning, VQA, transfer, physical deployment, and adaptive defense.

\runinhead{IR-VLMs and benchmarks.} Recent work has extended VLMs and multimodal large language models (MLLMs) to infrared semantic understanding. Infrared-LLaVA and IRGPT study model adaptation and infrared image-text data construction\cite{jiang2024infraredllava,cao2025irgpt}. Benchmarks such as RGB-Th-Bench and IF-Bench evaluate paired understanding and open-ended infrared reasoning\cite{moshtaghi2025rgbthbench,zhang2025ifbench}. These studies ask whether models can understand infrared content; our work asks whether their semantics remain faithful under physical perturbation. This distinction matters because infrared images encode shape, heat distribution, and low-texture body or object cues, allowing a local thermal patch to influence global category semantics if the visual encoder treats thermal contrast as reliable evidence. We therefore treat security evaluation as a complementary benchmark dimension for IR-VLMs.

\runinhead{Multimodal security and CLIP/VLM attacks.} As multimodal systems become stronger, robustness issues in multimodal reasoning have become increasingly clear. Recent studies on RGB VLMs and large vision-language models (LVLMs) show failures at both the image and cross-modal semantic levels, including visual jailbreaks, multimodal task attacks, transfer-based VLM attacks, physical-world LVLM patch attacks, large-scale targeted VLM attacks, and highly transferable attacks on CLIP representations\cite{qi2024visualjailbreak,yin2023vlattack,xie2025chainattack,liu2026spatial,zhang2025anyattack,huang2025xtransfer}. These methods show that aligned vision-language representations can be attacked without changing the language model itself. However, most operate in visible-light digital settings, optimize pixel-level or generator-produced perturbations, or rely on task-specific image-text objectives, and they do not need to fabricate a stable infrared artifact. Our work is positioned at the intersection of infrared physical security and multimodal semantic robustness: it studies a reusable thermal patch for IR-VLMs, attacks the clean-category manifold in visual representation space, and evaluates semantic vulnerability as a system-level property across classification, captioning, VQA, transfer, and physical deployment.

\runinhead{Derivative-free physical optimization.} Query-based optimization is widely used when gradients are unavailable or unreliable, including differential evolution, covariance-matrix adaptation, and score-based black-box attacks\cite{storn1997differential,hansen2001cmaes,ilyas2018blackbox,andriushchenko2020square}. For physical patches, the objective can be noisy because each candidate involves rendering, transformation, placement, and model evaluation. A derivative-free optimizer is therefore attractive, but the search space must be carefully constrained; otherwise, the optimizer may find high-frequency patterns that work digitally but are hard to fabricate or attach. The curved-grid parameterization in UCGP narrows the search to low-frequency curved structures; EOT samples capture-condition changes, whereas TPS models non-rigid patch deformation during optimization\cite{athalye2018eot,bookstein1989tps}. This combination is designed to make the search both query-efficient and deployment-aware.

\section{Method}
\label{sec:method}

\begin{figure*}[!t]
\centering
\includegraphics[width=0.96\textwidth]{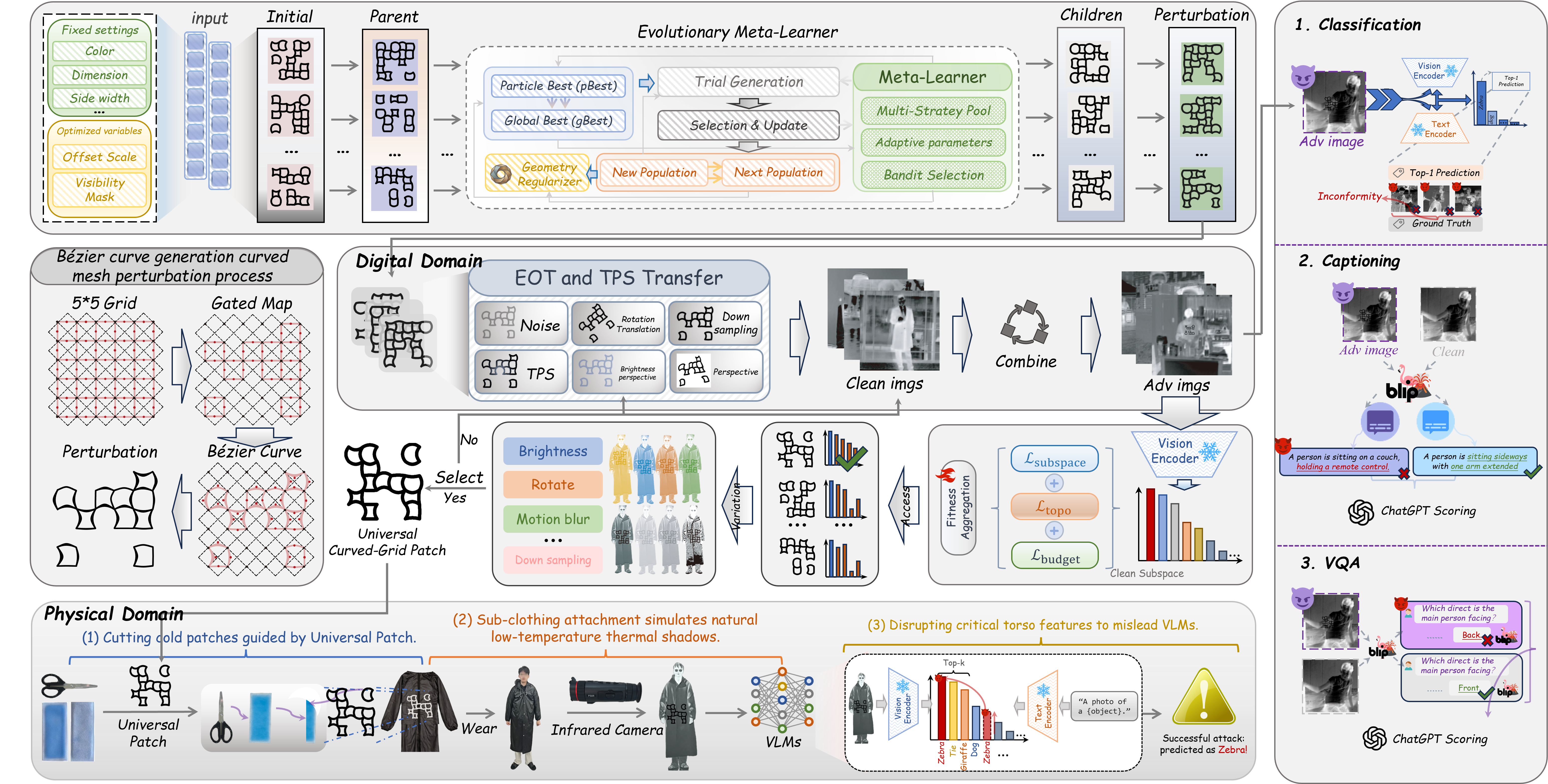}
\caption{Overall framework of \ourmethod{}. UCGP builds a clean-category reference structure, renders a universal Bézier curved-grid patch, and optimizes it with MetaDE using subspace, topology, and budget losses under EOT/TPS physical augmentations. The learned patch is reused across digital and physical IR-VLM evaluations reported here.}
\label{fig:pipeline}
\end{figure*}

\subsection{Problem Formulation}
\label{sec:problem-formulation}

An IR-VLM typically comprises a visual encoder $f(\cdot)$ that maps an infrared image $x$ to a normalized feature $z=f(x)$, and a language decoder or contrastive text encoder that produces open-ended outputs such as classification scores, captions, or VQA answers. Formally, given an infrared image set $\mathcal{D}=\{x_i\}_{i=1}^N$ and the corresponding target-region set $\{b_i\}$, our goal is to learn a single set of shared patch parameters $\theta$ and generate adversarial samples through a render-and-paste operator $R(\cdot,\cdot,\cdot)$:
\begin{equation}
x_i' = R(x_i, \theta, b_i),
\label{eq:render-and-paste}
\end{equation}
where $R$ instantiates the parameterized patch and overlays it on the target region specified by $b_i$. Equation~\eqref{eq:render-and-paste} formalizes the universal render-and-paste process. Because the same shared patch is reused across images and downstream VLM tasks, it must remain effective across samples, backgrounds, poses, and scene layouts.

Under this threat model, the attacker can query the IR-VLM through its forward interface and obtain classification, captioning, or VQA responses together with the normalized visual features $z=f(x)$ from the image encoder, but has no access to model gradients, parameters, or training data. We therefore refer to our setting as a representation-visible audit setting, not an output-only black-box setting. It is weaker than white-box optimization because model weights and gradients are unavailable, but stronger than commercial API-style output-only access because the continuous objective is built from visual representations. This interface is intended for controlled robustness auditing of models whose visual embeddings are exposed, logged, or recoverable; extending UCGP to strict output-only APIs is left as future work. During optimization, EOT samples final-image capture conditions and mild patch-level attachment variation, whereas TPS warps the rendered patch before it is pasted into the target region. Digital evaluations use rendered adversarial samples under this protocol, while physical testing is performed on real-world captured infrared videos. Fig.~\ref{fig:pipeline} gives an end-to-end view of UCGP through six steps: (1) clean infrared samples are passed through the IR-VLM visual encoder to build a clean-category reference structure; (2) a shared curved-grid patch is rendered from the patch parameters and pasted onto the target region of each image; (3) EOT capture sampling and TPS patch warping simulate imaging variation and local non-rigid deformation during candidate evaluation; (4) the resulting adversarial features are compared with the clean reference structure using subspace, topology, and budget losses; (5) MetaDE updates the shared patch parameters under this representation-level objective; and (6) the optimized patch is reused without per-image re-optimization for digital and physical evaluation across classification, captioning, and VQA. This six-step view emphasizes that UCGP is optimized in visual representation space, while its attack effect is ultimately measured through downstream semantic outputs such as predicted labels, generated captions, and VQA answers.

\subsection{Patch Generation}
\label{sec:patch-generation}

Directly optimizing a physical patch in pixel space creates a mismatch between digital search and deployable thermal artifacts. A pixel-wise patch has thousands of degrees of freedom even for a small region, and unconstrained derivative-free search can exploit isolated high-frequency pixels that improve digital scores but are unstable after fabrication, attachment, downsampling, and infrared capture. To address this, UCGP uses a curved-grid patch parameterization that constrains the patch to a low-dimensional geometric space composed of sparsely gated curved edges. Given a grid resolution of $G\times G$, let $\mathcal{E}_G$ denote the set of horizontal and vertical grid edges. The patch parameters consist of cell gates and edge-deformation variables:
\begin{equation}
\theta =
\{g_{u,v}\}_{u,v=1}^{G}
\cup
\{d_e\}_{e\in\mathcal{E}_G},
\label{eq:patch-parameters}
\end{equation}
where $g_{u,v}\in[0,1]$ controls whether cell $(u,v)$ contributes its boundary edges and $d_e$ controls the curvature of edge $e$. During rendering, the continuous gates are thresholded to select active cells; the boundary edges of active neighboring cells are shared and rendered once. For each selected edge with endpoints $p_0$ and $p_2$, $d_e$ shifts the quadratic Bézier control point away from the midpoint along the edge normal, with the maximum displacement bounded by the deformation ratio $\delta$ times the cell side length. The selected Bézier curves are rasterized into an anti-aliased soft mask $\alpha$, resized to the target region $b_i$, and blended with the chosen thermal color by $R(x_i,\theta,b_i)$.

\begin{figure*}[!t]
\centering
\includegraphics[width=0.72\textwidth]{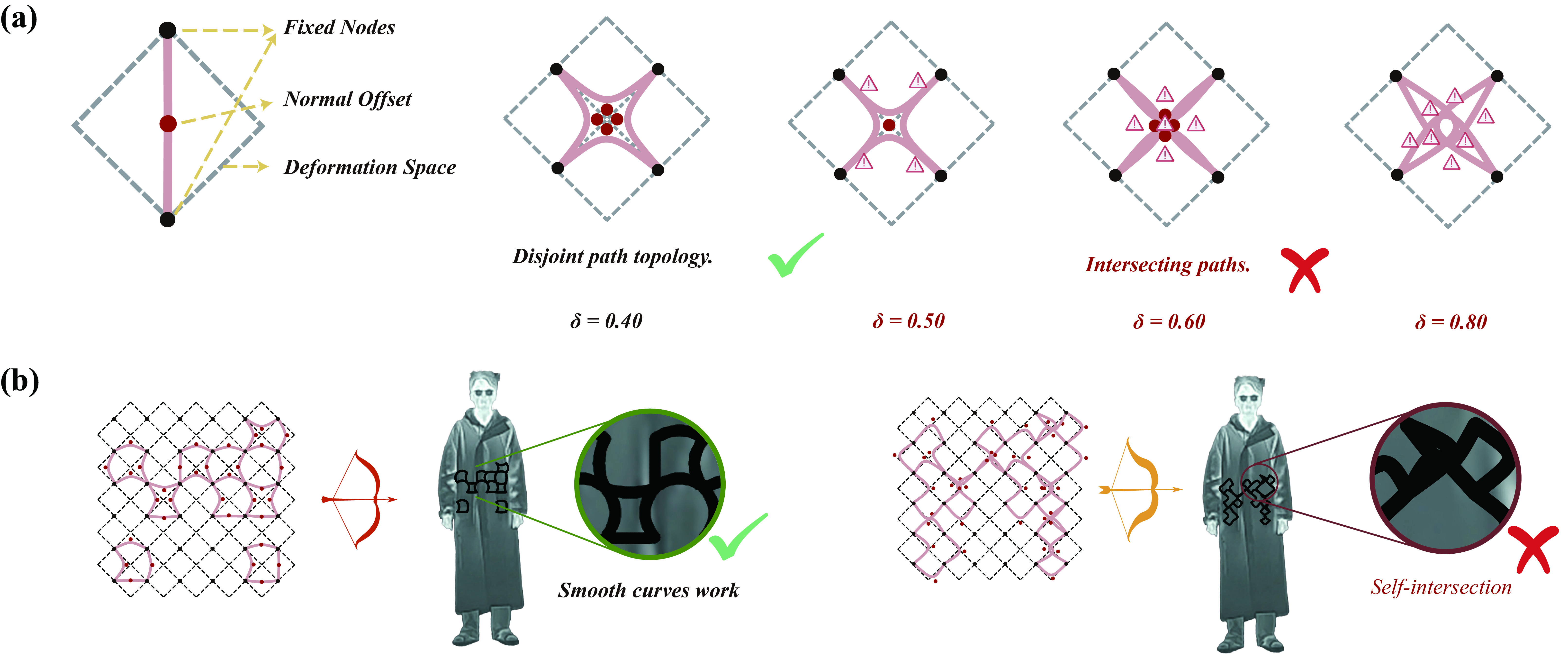}
\caption{Curved-grid parameterization examples. (a) Mesh topology under different deformation parameters. (b) Physical validation of smooth and self-intersecting structures during deployment.}
\label{fig:curved_grid}
\end{figure*}

This curved-grid parameterization injects a geometric inductive bias into the parameter space, steering the optimizer toward low-frequency structures that are easier to deploy, attach, and stably capture. Compared with pixel optimization, it reduces the search space from independent pixel intensities to cell gates and curve deformations, and it makes the thermal artifact interpretable as connected line segments rather than isolated noise. As shown in Fig.~\ref{fig:curved_grid}, when $\delta < 0.50$, the grid topology remains continuous and smooth; when $\delta \geq 0.50$, path intersections or self-intersections break contour continuity and create unstable enclosed regions. Physical deployment examples further show that continuous curves can be converted into attachable patches more reliably than topologically violated configurations. We therefore treat the curved-grid design as a deployability-oriented parameterization and validate its practical effect through the main comparison and parameter-sensitivity analyses.

To prevent the patch from overfitting to an idealized digital environment, we evaluate every candidate under two distinct physical robustness augmentations rather than adding a separate attack objective. TPS is used for local patch deformation, while EOT is used for final-image capture variation. This follows standard physical infrared attack practice: the rendered perturbation is first deformed to mimic cloth folding and local non-rigid motion, and the pasted adversarial image is then evaluated under sampled capture conditions~\cite{tiliwalidi2024infraredgrid,athalye2018eot,bookstein1989tps}. Let $P_\theta=(\alpha_\theta,c_\theta)$ denote the rendered soft patch mask and patch appearance in normalized patch coordinates. For each image-region pair $(x_i,b_i)$, the deterministic rendered sample without robustness sampling is
\begin{equation}
\bar{x}_i(\theta)=R(x_i,P_\theta,b_i).
\label{eq:base-render}
\end{equation}
We sample a composite capture transformation $t{\sim}\mathcal{T}$. The transformation family contains mild patch-level scale, translation, rotation, and shear jitter before regional pasting, together with final-image affine, brightness/contrast, and sensor-noise perturbations. Following Wei et al.~\cite{wei2023unified}, who synthesize multi-angle samples using body-keypoint-assisted homographies, we approximate left/right viewpoint changes by ROI-level patch rotation and bidirectional shear. Under EOT~\cite{athalye2018eot}, these view-like deformations expose each candidate to limited nonfrontal renderings during optimization.

Because the patch is attached to clothing and can undergo local non-rigid deformation, we further model cloth deformation with a TPS warp~\cite{bookstein1989tps}. Let $\mathcal{C}=\{c_m\}_{m=1}^{K_c}$ be control points on the patch/region-of-interest (ROI) plane and let $\omega=\{\Delta c_m\}_{m=1}^{K_c}$ be sampled control-point displacements, giving target points $\hat{c}_m=c_m+\Delta c_m$. The induced TPS mapping is
\begin{equation}
\begin{aligned}
\phi_{\omega}(p)
&= a_0 + A p
+ \sum_{m=1}^{K_c} w_m U(\lVert p-c_m\rVert_2),\\
U(r)&=r^2\log(r^2+\epsilon),
\end{aligned}
\label{eq:tps-warp}
\end{equation}
where $p$ is a patch-plane coordinate; $a_0\in\mathbb{R}^2$ is the translation vector, $A\in\mathbb{R}^{2\times2}$ is the affine matrix, and $w_m\in\mathbb{R}^2$ is the radial-basis weight associated with control point $c_m$. The coefficients $\{a_0,A,w_m\}$ are solved from $\phi_{\omega}(c_m)=\hat{c}_m$ with the standard TPS affine constraints, and $\epsilon>0$ is a small constant for numerical stability. Image values after TPS are obtained by bilinear sampling under this deformation field. The TPS-warped patch is
\begin{equation}
P_{\theta,\omega}(p)=P_\theta\!\left(\phi_{\omega}^{-1}(p)\right),
\label{eq:tps-patch}
\end{equation}
where the inverse warp indicates backward sampling from the original rendered patch. Thus, $\phi_\omega$ acts on the rendered soft patch mask and patch appearance in the patch/ROI plane before regional pasting, while $t$ is applied to the final adversarial image to simulate capture variation. The physically augmented adversarial image is therefore
\begin{equation}
\tilde{x}_i(\theta;t,\omega)
=t\!\left(R\!\left(x_i,P_{\theta,\omega},b_i\right)\right).
\label{eq:augmented-render}
\end{equation}
The robust optimization target is
\begin{equation}
\begin{aligned}
\theta^{\star}
&=
\arg\min_{\theta}\;
\mathbb{E}_{(x_i,b_i)\sim\mathcal{D}}
\mathbb{E}_{t{\sim}\mathcal{T},\,\omega{\sim}\Omega}\\
&\quad
\left[
\mathcal{L}\!\left(f(\tilde{x}_i(\theta;t,\omega))\right)
\right],
\end{aligned}
\label{eq:robust-objective}
\end{equation}
where $f(\cdot)$ is the visual encoder or downstream IR-VLM interface and $\mathcal{L}$ denotes the representation-level objective defined below. Equations~\eqref{eq:augmented-render} and~\eqref{eq:robust-objective} respectively define one physically augmented sample and the robust objective over such samples. In practice, the nested expectation is approximated by Monte Carlo samples during candidate evaluation. The local patch-level affine/shear jitter is included in $\mathcal{T}$ as a limited multi-scale and multi-view approximation, keeping the implementation lightweight and consistent with the EOT formulation used here.

\subsection{Objective Design}
\label{sec:representation-objective}

The core idea of this work is to disrupt the clean-category manifold rather than directly optimizing a particular class score. The framework is category-agnostic: for any target category, we construct a clean-category reference set $\mathcal{D}_{\text{clean}}$ from samples that the model stably recognizes as belonging to that category under clean input; in our main experiments, this is instantiated as person. For each sample we extract the normalized visual feature $z_i = f(x_i)$ and compute the mean:
\begin{equation}
\mu = \frac{1}{|\mathcal{D}_{\text{clean}}|}\sum_i z_i.
\label{eq:clean-mean}
\end{equation}
We then perform principal component analysis (PCA) on the centered features, implemented via singular value decomposition (SVD), to obtain the top-$k$ principal directions forming the subspace basis $U_k$, and further estimate the kernel scale required by the graph-based Kullback--Leibler (Graph-KL) divergence from neighborhood relations among the clean features. The clean reference mean is defined in Eq.~\eqref{eq:clean-mean}. The columns of $U_k$ summarize dominant clean intra-category variations, such as pose, scale, and thermal contrast. Therefore, movement within this subspace is treated as normal category variation, while movement orthogonal to it indicates departure from the learned target-category structure.


This reference structure represents target semantics as a stable statistical structure on a low-dimensional subspace and a local neighborhood graph, rather than as a specific text label. The loss is applied to the visual representation, while text-side behavior is used for downstream evaluation. Because the structure is estimated from cross-sample statistics, optimization targets one shared patch parameter $\theta$ instead of learning per-image perturbations. The overall objective comprises the following three terms:
\begin{equation}
\mathcal{L} = \mathcal{L}_{\text{subspace}}
+ \lambda_{\text{topo}} \mathcal{L}_{\text{topo}}
+ \lambda_{\text{budget}} \mathcal{L}_{\text{budget}},
\label{eq:total-loss}
\end{equation}
where the three terms in Eq.~\eqref{eq:total-loss} correspond to subspace departure, topology disruption, and local appearance regularization, respectively. The local appearance regularizer constrains thermal statistics, boundary smoothness, and the spatial extent of the patch; it does not aim to make the patch invisible in infrared imagery. The three terms jointly act on the shared patch parameter $\theta$, thereby learning a universal patch that can be reused across images.

The first term captures subspace departure. For an adversarial sample $x_i'$, we extract its visual feature $z_i' = f(x_i')$ and define its orthogonal deviation energy relative to the clean-category principal subspace as
\begin{equation}
r_i = \left\|(I-U_kU_k^\top)(z_i' - \mu)\right\|_2^2.
\label{eq:subspace-deviation}
\end{equation}
To enable the shared patch to stably disrupt the clean-category representation structure, we maximize the average deviation energy across all samples in the batch:
\begin{equation}
\mathcal{L}_{\text{subspace}} = - \frac{1}{B}\sum_{i=1}^{B} r_i,
\label{eq:subspace-loss}
\end{equation}
where $B$ is the mini-batch or candidate evaluation subset size. Thus, Eqs.~\eqref{eq:subspace-deviation} and~\eqref{eq:subspace-loss} define the per-sample departure and its batch-level objective. Because the optimization objective is minimization, the negative sign encourages increasing the subspace deviation of the samples.

The second term captures topology disruption. Because subspace departure can be dominated by a few large feature movements, we also compare whether local neighborhoods inside the target-category batch are preserved. For clean features $\{z_i\}$ and adversarial features $\{z_i'\}$, we compute row-normalized Gaussian affinities:
\begin{equation}
\begin{aligned}
P_{ij}
&=
\frac{\exp(-\lVert z_i-z_j\rVert_2^2/\sigma)}
{\sum_{\ell\ne i}\exp(-\lVert z_i-z_\ell\rVert_2^2/\sigma)},\\
Q_{ij}
&=
\frac{\exp(-\lVert z_i'-z_j'\rVert_2^2/\sigma)}
{\sum_{\ell\ne i}\exp(-\lVert z_i'-z_\ell'\rVert_2^2/\sigma)},
\end{aligned}
\label{eq:graph-affinity}
\end{equation}
where self-affinities are excluded and $\sigma$ is estimated from the median pairwise squared distance of clean reference features. The affinities in Eq.~\eqref{eq:graph-affinity} are then used to maximize the Graph-KL divergence between the clean and adversarial neighborhood distributions~\cite{hinton2002sne,vandermaaten2008tsne}:
\begin{equation}
\mathcal{L}_{\text{topo}} =
- \frac{1}{B}\sum_{i}\sum_{j\ne i}
P_{ij}\log\frac{P_{ij}}{Q_{ij}}.
\label{eq:topology-loss}
\end{equation}
Minimizing the topology term in Eq.~\eqref{eq:topology-loss} is equivalent to maximizing the KL divergence between the neighborhood graphs, thereby disrupting the local structural stability of the clean-category manifold. Compared with a simple mean shift, this term emphasizes ``structural misalignment.''
The third term regularizes the local appearance, boundary smoothness, and spatial extent of the patch:
\begin{equation}
\mathcal{L}_{\text{budget}}
=
\mathcal{L}_{\text{therm}}
 +
\mathcal{L}_{\text{edge}}
+
\mathcal{L}_{\text{area}}.
\label{eq:budget-loss}
\end{equation}
Let $\alpha$ denote the soft activation mask of the patch region and $\beta$ the surrounding background ring obtained by morphological dilation. The thermal-statistics term penalizes distributional mismatch between the patch and its local context:
\begin{equation}
\mathcal{L}_{\text{therm}}
= (\mu_\alpha - \mu_\beta)^{2}
+ (\sigma_\alpha - \sigma_\beta)^{2},
\label{eq:thermal-loss}
\end{equation}
where $\mu_\alpha,\sigma_\alpha$ and $\mu_\beta,\sigma_\beta$ are the intensity mean and standard deviation weighted by $\alpha$ and $\beta$, respectively. The thermal term in Eq.~\eqref{eq:thermal-loss} is complemented by a boundary-smoothness term that suppresses high-frequency artifacts along the patch edge:
\begin{equation}
\mathcal{L}_{\text{edge}}
= \max\bigl(0,\;
\bar{g}_{\partial\alpha} - \bar{g}_\beta\bigr),
\label{eq:edge-loss}
\end{equation}
where $\bar{g}_{\partial\alpha}$ is the average image-gradient magnitude along the activation boundary and $\bar{g}_\beta$ is the average gradient in the background ring. The area term in Eq.~\eqref{eq:area-loss} constrains effective coverage:
\begin{equation}
\mathcal{L}_{\text{area}}
= \bar{\alpha}^{\,2},
\quad 
\bar{\alpha}=\frac{\sum\alpha}{A_{\text{bbox}}},
\label{eq:area-loss}
\end{equation}
where $A_{\text{bbox}}$ is the bounding-box area. The complete regularizer in Eq.~\eqref{eq:budget-loss} combines the three terms in Eqs.~\eqref{eq:thermal-loss}, \eqref{eq:edge-loss}, and~\eqref{eq:area-loss}. Together, these terms improve local appearance consistency and constrain the patch area; infrared invisibility is neither assumed nor optimized.

\subsection{Optimization Procedure}
\label{sec:metade-optimization}

Evaluating a candidate patch involves soft rendering, TPS patch warping, regional pasting, EOT capture sampling, and IR-VLM forward inference, making the objective function non-differentiable with respect to $\theta$. Meanwhile, the UCGP parameterization restricts the parameter space to a low-dimensional structured form where each evaluation is expensive and noisy. Given these problem characteristics, we adopt MetaDE\cite{chen2026metade} for derivative-free search to balance exploration with stable convergence under a limited query budget.

In our setting, the fitness function approximates the robust objective over the current evaluation subset and sampled physical transformations:
\begin{equation}
\begin{aligned}
F(\theta)
&= \frac{1}{K}\sum_{s=1}^{K}\mathcal{J}_s(\theta),\\
\mathcal{J}_s(\theta)
&= \Lsub(\tilde{\mathcal{Z}}_s;\mu,U_k)\\
&\quad + \lambda_{\text{topo}}
\Ltop(\tilde{\mathcal{Z}}_s,\mathcal{Z}_{\mathcal{B}})\\
&\quad + \lambda_{\text{budget}}
\Lbud(\tilde{\mathcal{B}}_s,\mathcal{B}).
\end{aligned}
\label{eq:metade-fitness}
\end{equation}
where each transformed sample is
\begin{equation*}
\tilde{x}_{i,s}
=t_s\!\left(R\!\left(x_i,P_{\theta,\omega_s},b_i\right)\right).
\end{equation*}
Here, $\tilde{\mathcal{B}}_s(\theta)$ is the transformed evaluation subset, $\tilde{\mathcal{Z}}_s=\{f(\tilde{x}_{i,s})\}_{(x_i,b_i)\in\mathcal{B}}$ is the corresponding feature set, $\mathcal{Z}_{\mathcal{B}}=\{f(x_i)\}_{(x_i,b_i)\in\mathcal{B}}$ is the clean reference feature set, and $K$ is the number of Monte Carlo samples, each containing one EOT capture transform and one TPS patch warp. The two samples act in different domains: $\omega_s$ deforms the rendered patch in patch/ROI coordinates, whereas $t_s$ transforms the composed adversarial image. Thus, $F(\theta)$ is not a new objective; it is the Monte Carlo estimator of the robust loss used to compare patch candidates. The index $s$ represents sampled physical conditions, while $\mathcal{B}$ represents the current image subset. The subspace and budget terms use the batch averages defined above, while the topology term is computed on the clean and adversarial neighborhood distributions of the current transformed subset. Through iterative optimization, MetaDE learns a shared patch under practical conditions that stably disrupts target-category representations. The overall training procedure is shown in Algorithm~\ref{alg:ucgp_optimization}, which encompasses clean-category reference structure precomputation, UCGP patch rendering, physical robustness evaluation, and MetaDE evolutionary updates.

\begin{algorithm}[tb]
\caption{Optimization pseudocode of \ourmethod{}.}
\label{alg:ucgp_optimization}
\small
\begin{algorithmic}[1]
\Require Target sample set $\mathcal{D}$ with regions $\{b_i\}$, encoder $f$, population size $P$, generations $T$
\Ensure Universal patch $\theta^{\star}$
\State Build local target-region samples from $(\mathcal{D}, \{b_i\})$
\State Compute clean statistics $(\mu,U_k,\sigma)$ from $z_i=f(x_i)$
\State Initialize MetaDE population $\{\theta_j^{(0)}\}_{j=1}^{P}$
\State Set $\theta^{\star} \gets \arg\min_{j} F(\theta_j^{(0)})$
\For{$t \gets 0$ to $T-1$}
  \State Sample evaluation subset $\mathcal{B}_t \subseteq \mathcal{D}$
  \State Update MetaDE evolution strategy
  \For{each candidate $\theta_j^{(t)}$}
    \State Generate trial $\tilde{\theta}_j$ by mutation and crossover
    \For{each sample $(x,b) \in \mathcal{B}_t$}
      \State Sample an EOT capture transform $t_s$ and a TPS patch warp $\omega_s$
      \State Render $P_{\tilde{\theta}_j}$, warp it as $P_{\tilde{\theta}_j,\omega_s}$, and form $\tilde{x}_s=t_s(R(x,P_{\tilde{\theta}_j,\omega_s},b))$
      \State Extract $z'_s=f(\tilde{x}_s)$ for aggregate computation of $\Lsub$, $\Ltop$, and $\Lbud$
    \EndFor
    \State Evaluate $F(\tilde{\theta}_j)$ per the fitness function above
    \If{$F(\tilde{\theta}_j) < F(\theta_j^{(t)})$}
      \State Accept $\tilde{\theta}_j$
    \EndIf
  \EndFor
  \State Update best candidate $\theta^{\star}$
  \If{stagnation occurs}
    \State Refresh part of the population
  \EndIf
\EndFor
\State \Return $\theta^{\star}$
\end{algorithmic}
\end{algorithm} 
Throughout optimization and deployment, we maintain and reuse a single patch parameter vector $\theta$ across images, local target layouts, and datasets without per-image optimization. This is the universal patch setting studied in this paper.

\section{Experiments}
\label{sec:experiments}

\subsection{Experimental Setup}
\label{sec:experimental-setup}
\runinhead{Datasets.} Except for transfer experiments, all main experiments are conducted on Infrared-COCO for classification, captioning, and VQA\cite{jiang2024infraredllava,lin2014microsoft}. Under our category-agnostic framework, the primary experimental instantiation uses person as the target category: we keep valid images with person-class probability above 0.50, then sample 300 images for the three main tasks. This shared image pool keeps the main cross-task comparison consistent, so the observed differences are driven by task outputs rather than by different evaluation subsets. Defense evaluation and auxiliary analyses use the same retrieval and filtering protocol unless otherwise specified. For cross-category extension, we rebuild the clean-category reference set for each target category and run the same UCGP optimization pipeline without category-specific redesign.

We use person as the main target category because it is both practically relevant for infrared sensing and sufficiently represented across the evaluated datasets. This choice also creates a challenging setting: person instances vary greatly in pose, distance, clothing, occlusion, and thermal contrast. A universal patch must therefore work across different body shapes and local attachment contexts rather than overfitting to a small set of near-duplicate crops. In all main experiments, the clean images are kept fixed across methods and tasks whenever possible, which makes differences between attacks easier to attribute to the patch design and objective rather than to sampling noise.

For transfer evaluation, we use LSOTB-TIR~\cite{liu2020lsotbtir}, LLVIP~\cite{jia2021llvip}, M3FD~\cite{liu2022tardal}, and FLIR ADAS v1.3~\cite{fliradasv13}. These datasets provide complementary domain shifts: LSOTB-TIR contains tracking sequences with scale and motion variation, LLVIP emphasizes low-light visible-infrared scenes, M3FD contains multi-spectral detection scenes with complex backgrounds, and FLIR ADAS v1.3 provides thermal driving scenarios. This diversity makes transfer more informative than a same-distribution split, since a patch that transfers across these datasets is less likely to exploit only dataset-specific backgrounds or annotations. We therefore report attack success rate (ASR) together with the conditional score-movement descriptors defined below.

\runinhead{Models and Fine-Tuning Protocol.} Following Infrared-LLaVA\cite{jiang2024infraredllava}, we build both classification models and generative IR-VLMs on a unified infrared semantic corpus. Infrared-COCO provides about 17k infrared images synthesized from Microsoft Common Objects in Context (MS-COCO)\cite{lin2014microsoft} and 12k matched QA pairs\cite{jiang2024infraredllava}. We evaluate LLaVA-1.5\cite{liu2024llava15}, LLaVA-1.6\cite{liu2024llavanext}, OpenFlamingo\cite{awadalla2023openflamingo}, BLIP-2\cite{li2023blip2}, and InstructBLIP\cite{dai2023instructblip} for captioning/VQA, and OpenAI CLIP ViT-L/14\cite{radford2021clip}, OpenCLIP ViT-B/16\cite{cherti2023openclip}, Meta-CLIP ViT-L/14\cite{xu2023metaclip}, and EVA-CLIP ViT-G/14\cite{sun2023evaclip} for classification. This model suite spans both contrastive classification and generative semantic interfaces.

All models start from natural-image-pretrained weights, freeze the backbone, and use low-rank adaptation (LoRA)\cite{hu2022lora}. For captioning and VQA, all image-question-answer pairs in Infrared-COCO are used for next-token prediction fine-tuning. For classification, we select the question-answer pairs applicable to category prediction, retain the answer as the class label, and optimize the contrastive encoders with Information Noise-Contrastive Estimation (InfoNCE)\cite{oord2018cpc}. Training stops when the validation loss no longer decreases. Unless otherwise specified, later classification experiments use OpenAI CLIP ViT-L/14 as both the surrogate and evaluation model. Classification follows the 80-category MS-COCO label space, where predicting any non-person class for a person sample is counted as a successful attack\cite{lin2014microsoft}.

This model protocol is intended to separate two factors: the infrared adaptation of the visual representation and the downstream language interface. Contrastive encoders provide a controlled classification setting where score changes can be directly measured, while LLaVA, OpenFlamingo, BLIP-2, and InstructBLIP test whether the same visual perturbation affects generated language. We do not tune a different patch for each downstream task in the main comparison. Instead, the learned universal patch is evaluated across interfaces, which makes the result more conservative than a task-specific attack and better matches the claim that the vulnerability lies in shared visual semantics.

\runinhead{Evaluation Metrics.} We report attack effectiveness using task-specific success rates and clean-to-adversarial score changes. We additionally report image naturalness as a perceptual sanity check. Let $X=\{x_i\}_{i=1}^{N}$ and $X'=\{x_i'\}_{i=1}^{N}$ denote paired clean and adversarial evaluation sets, and let $y^\star$ denote the evaluated target category, instantiated as person in the main experiments. Let $s_c(x)$ denote the model-specific pre-softmax CLIP logit
for class $c$, computed from the scaled cosine similarity between
the normalized image and class-text embeddings. The corresponding
class probability is obtained by applying softmax to all class
logits. For classification, the untargeted attack success rate (ASR) is defined as
\begin{equation}
\mathrm{ASR}
= \frac{1}{N}\sum_{i=1}^{N}
\mathbf{1}\!\left[\arg\max_{c} s_c(x_i') \ne y^\star\right].
\end{equation}
For captioning and VQA, following the large language model (LLM)-as-a-judge protocol in~\cite{liu2025lighting}, we use fixed judge templates: captioning is scored by semantic consistency and normalized to a 0--100 Consistency Score from the 0--30 judge total, while VQA is scored by binary answer correctness and averaged as a 0--100 Correctness Score. The down-arrow values in the captioning and VQA tables denote the absolute score decrease from the clean output to the adversarial output; larger decreases indicate stronger semantic degradation.

For transfer, we additionally report two conditional score-movement descriptors based on the classification score $s_c(\cdot)$ for class $c$. Let $\mathcal{S}=\{i:\arg\max_c s_c(x_i')\neq y^\star\}$ denote the set of successfully attacked samples. For each $i\in\mathcal{S}$, we select the strongest promoted competing class and define
\begin{align}
c_i^\ast
&= \arg\max_{c\neq y^\star} s_c(x_i'), \\
\Delta s_{\mathrm{prom}}
&= \frac{1}{|\mathcal{S}|}\sum_{i\in\mathcal{S}}
\left(s_{c_i^\ast}(x_i')-s_{c_i^\ast}(x_i)\right), \\
M_{\mathrm{adv}}
&= \frac{1}{|\mathcal{S}|}\sum_{i\in\mathcal{S}}
\left(s_{c_i^\ast}(x_i')-s_{y^\star}(x_i')\right).
\end{align}
ASR measures the frequency of successful attacks. Conditional on success, $\Delta s_{\mathrm{prom}}$ characterizes promotion of the winning competing class relative to its clean score, and $M_{\mathrm{adv}}$ characterizes its final margin over the original class. These score-movement descriptors complement, rather than replace, ASR. For Table~\ref{tab:cls_nat}, we additionally evaluate image naturalness, denoted Nat., using a GPT-5-based LLM-as-a-judge protocol on a 5-point scale; the scoring criteria, constrained output format, and prompt templates used for the judge are provided in the supplementary material.

For LLM-as-a-judge evaluation, we use the same judge setting across methods and compare clean and adversarial outputs under the same rubric, since generative models differ in clean output quality. The prompt is fixed across clean and adversarial outputs, so the comparison measures changes induced by the visual patch rather than changes in evaluation wording. The naturalness score is used only as a perceptual sanity check for local appearance; it is not optimized by our method and should not be interpreted as infrared invisibility. Larger-scale human validation and judge-sensitivity analysis remain useful future directions.

\runinhead{Baselines.} Under the experimental settings above, we compare against three representative physically deployable infrared baselines: HCB, a block-based thermal-patch baseline\cite{hu2023infraredblocks}; AdvGrid\cite{tiliwalidi2024infraredgrid}; and QR Code\cite{zhu2022invisibleclothing}. Although originally designed for infrared detection or related physical attacks, they are adapted to the unified IR-VLM threat model, loss, and evaluation protocol used here, with the same target regions, evaluation images, and downstream models as \ourmethod{}. The baselines represent different structural priors: QR Code provides a high-contrast regular pattern, HCB uses block-like thermal perturbations, and AdvGrid uses a grid prior close to our structured patch family. Recent RGB/CLIP attacks are valuable indirect baselines, but their default forms are often digital, visible-spectrum, or prompt-dependent; a fair physical infrared comparison would require separate adaptation to thermal capture, fabrication, region constraints, and the same evaluation protocol.

\begin{figure}[!t]
\centering
\includegraphics[width=0.98\columnwidth]{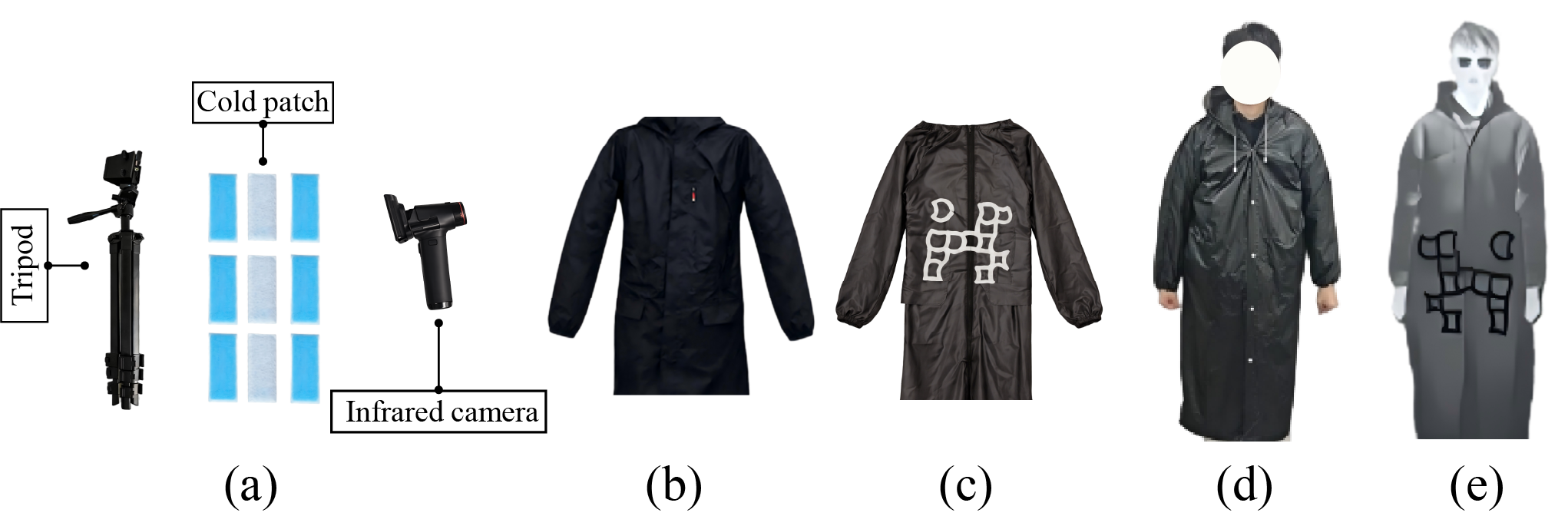}
\caption{Experimental setup. (a) Tripod, cold patches, and infrared camera. (b) Visible-spectrum image of the outer garment. (c) Visible-spectrum image of the inner garment with the patch attached. (d) Visible-spectrum image of the wearer after deployment. (e) Infrared image of the wearer.}
\label{fig:physical_setup}
\end{figure}

\runinhead{Implementation Details.} For the curved-grid parameterization, we use grid dimension $G=5$, a square patch with side length equal to one-quarter of the target height, and curvature-control parameter $\delta=0.40$ to preserve smooth, non-self-intersecting curves during deployment. MetaDE uses population size $P=50$, $T=100$ generations, and a fixed query budget. The grid-line width is 0.20 times the cell-edge length, and the patch is black by default. Loss weights are $\lambda_{\text{topo}}=0.12$ and $\lambda_{\text{budget}}=0.03$. All adversarial experiments are executed on a workstation equipped with eight NVIDIA RTX 4090 graphics processing units (GPUs); unless otherwise specified, each optimization/evaluation worker uses one GPU and independent jobs are parallelized across GPUs.

Hyperparameters are selected once and reused across the main experiments, avoiding per-setting retuning. The physical experiments use the same algorithmic UCGP configuration as the digital evaluation, including the curved-grid parameterization, patch scale, patch color, line width, loss weights, and optimized patch parameter vector; no physical-specific re-optimization or hyperparameter retuning is performed before deployment. The default patch scale is large enough to affect the visual representation but small enough to remain attachable to common person regions. The topology weight and grid-line width balance attack strength against physical stability: overly sparse patterns have weak thermal influence, while overly dense or self-intersecting patterns are difficult to deploy reliably.

For reproducibility, patch parameters, model identifiers, preprocessing configurations, judge prompts, and physical capture metadata are logged for each run. The implementation is organized around a renderer, an optimizer, and task-specific evaluation adapters, so the same patch candidate can be evaluated under classification, captioning, and VQA without changing the optimization code.

\runinhead{Experimental Devices.} For physical evaluation, the setup is shown in Fig.~\ref{fig:physical_setup}. Physical experiments use an InfiRay XL19V2 infrared camera ($384{\times}288$ resolution, ${<}\,18$\,mK sensitivity) mounted on a tripod. We use cold patches that maintain $24^{\circ}$C for 10 hours and produce strong infrared perturbations. In our setup, the patch is attached to the inner garment and covered by the outer garment, while remaining clearly visible in infrared images.

During physical capture, we vary distance and viewing angle to test whether the optimized patch remains effective beyond a single carefully aligned shot. The indoor sequence covers 6.2--11.0 m, whereas the outdoor sequence covers 7.8--12.6 m because the open outdoor space allows and motivates longer-range testing. Following common infrared physical-attack protocols~\cite{tiliwalidi2024infraredgrid}, we record continuous indoor and outdoor infrared videos rather than isolated photographs; the submitted supplementary videos cover approximately 80--90 seconds per scene. We uniformly sample valid video intervals at 3 frames per second to reduce temporal redundancy while keeping enough frames for stable statistics. This protocol is needed because infrared deployment is sensitive to apparent patch size, body pose, cloth deformation, and background temperature, and a patch that works only in a close-range frontal view would be less convincing as a physical attack.

\begin{table*}[!t]
\centering
\caption{Classification ASR and naturalness results. Nat. is the GPT-5-based 5-point naturalness score, where higher is more natural. Cited baselines are adapted to our unified IR-VLM region, loss, and evaluation protocol; dashes mark metrics not applicable to clean inputs.}
\label{tab:cls_nat}
\footnotesize
\setlength{\tabcolsep}{5pt}
\renewcommand{\arraystretch}{1.05}
\resizebox{0.94\textwidth}{!}{%
\begin{tabular}{L{3.2cm} *{8}{C{1.4cm}}}
\toprule
\multirow{2}{*}{\raisebox{-0.4ex}{\textbf{Method}}} &
\multicolumn{2}{c}{\textbf{OpenAI CLIP ViT-L/14}} &
\multicolumn{2}{c}{\textbf{OpenCLIP ViT-B/16}} &
\multicolumn{2}{c}{\textbf{Meta-CLIP ViT-L/14}} &
\multicolumn{2}{c}{\textbf{EVA-CLIP ViT-G/14}} \\
\cmidrule(lr){2-3}\cmidrule(lr){4-5}\cmidrule(lr){6-7}\cmidrule(lr){8-9}
& \textbf{ASR} & \textbf{Nat.} & \textbf{ASR} & \textbf{Nat.} & \textbf{ASR} & \textbf{Nat.} & \textbf{ASR} & \textbf{Nat.} \\
\midrule
Clean & -- & 3.26 & -- & 3.26 & -- & 3.26 & -- & 3.26 \\
AdvGrid~\cite{tiliwalidi2024infraredgrid}
  & 19.33 & \second{2.21}\dec{1.05}
  & 25.00 & \second{2.15}\dec{1.11}
  & 19.67 & \best{2.26}\dec{1.00}
  & \second{27.67} & \second{2.24}\dec{1.02} \\
HCB~\cite{hu2023infraredblocks}
  & 20.33 & 2.08\dec{1.18}
  & \second{34.00} & 2.04\dec{1.22}
  & \second{20.67} & 2.07\dec{1.19}
  & 20.00 & 2.16\dec{1.10} \\
QR Code~\cite{zhu2022invisibleclothing}
  & \second{22.67} & 2.15\dec{1.11}
  & 20.33 & 2.10\dec{1.16}
  & 19.00 & 2.12\dec{1.14}
  & 17.67 & 2.06\dec{1.20} \\
\rowcolor{OursGray}
\textbf{\ourmethod{}}
  & \best{94.33} & \best{2.33}\dec{0.93}
  & \best{61.67} & \best{2.30}\dec{0.96}
  & \best{34.33} & \second{2.25}\dec{1.01}
  & \best{48.00} & \best{2.37}\dec{0.89} \\
\bottomrule
\end{tabular}
}
\end{table*}

\subsection{Comparison with Baselines}
\label{sec:comparison-with-other-methods}
\begin{figure*}[t]
\centering
\includegraphics[width=0.88\textwidth]{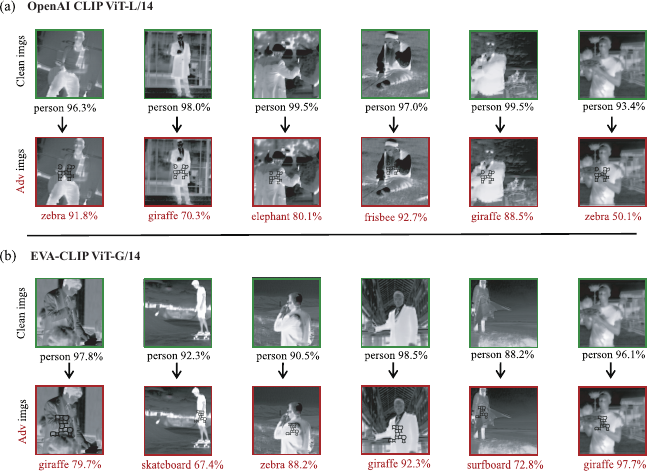}
\caption{UCGP-generated adversarial samples and their
classification results. The labels below each image show the
top-1 predicted class and its softmax probability.
(a) Results on OpenAI CLIP ViT-L/14.
(b) Results on EVA-CLIP ViT-G/14.}
\label{fig:classification_examples}
\end{figure*}

\runinhead{Classification Results.} Table~\ref{tab:cls_nat} reports classification ASR and image naturalness, denoted Nat., across four CLIP backbones. \ourmethod{} achieves the best or clearly leading ASR on all four backbones while retaining comparable local appearance scores. On OpenAI CLIP ViT-L/14, it reaches 94.33\% ASR, far above the strongest baseline at 22.67\%; on Meta-CLIP ViT-L/14, the most robust backbone in this table, it still reaches 34.33\% ASR versus 20.67\% for the best baseline. The backbone differences show that representation robustness is uneven: OpenAI CLIP ViT-L/14 is the most vulnerable, Meta-CLIP ViT-L/14 is the most robust, and EVA-CLIP ViT-G/14 and OpenCLIP ViT-B/16 fall in between. Across all backbones, the large ASR gaps and comparable local appearance scores indicate that disrupting the clean-category subspace is more effective than the structured patch baselines considered here. Fig.~\ref{fig:classification_examples} shows that the same patch changes predicted semantics on both OpenAI CLIP and EVA-CLIP, so the effect is not tied to a single classifier head.

The classification results also clarify why a representation-level objective is useful. QR Code, HCB, and AdvGrid introduce visible or thermal structures that can disturb local evidence, but their ASR remains modest because the global person representation often stays intact. In contrast, \ourmethod{} is optimized to move features away from the clean-category manifold, so the perturbation does not need to imitate a particular non-person object. This helps explain why the attack remains effective across different CLIP-family encoders despite their different pretraining data and model scales. The naturalness scores serve only as a perceptual sanity check, suggesting that the performance gain is not solely caused by a more visually disruptive pattern.

\begin{table*}[t]
\centering
\caption{Captioning results. Scores are GPT-5-judged semantic consistency scores on a 0--100 scale; down arrows denote absolute drops from clean outputs. Adapted baselines are defined and cited in Table~\ref{tab:cls_nat}.}
\label{tab:captioning_results}
\footnotesize
\setlength{\tabcolsep}{3.8pt}
\renewcommand{\arraystretch}{1.08}
\begin{tabular}{C{1.85cm} C{3.95cm} *{4}{C{1.52cm}} G{1.68cm}}
\toprule
\textbf{Image Encoder} & \textbf{Model} & \textbf{Clean} & \textbf{QR Code} & \textbf{HCB} & \textbf{AdvGrid} & \oursheader \\
\midrule
\multirow{4}{*}{\makecell[c]{OpenAI CLIP\\ViT-L/14}}
& LLaVA-1.5 (7B) & 66.03 & 61.59\dec{4.44} & 61.99\dec{4.04} & \second{61.00}\dec{5.03} & \best{48.79}\dec{17.24} \\
& LLaVA-1.6 (13B) & 68.01 & \second{61.09}\dec{6.92} & 63.94\dec{4.07} & 63.65\dec{4.36} & \best{52.77}\dec{15.24} \\
& OpenFlamingo (3B) & 63.17 & 58.37\dec{4.80} & \second{51.37}\dec{11.80} & 54.93\dec{8.24} & \best{43.77}\dec{19.40} \\
& BLIP-2 FlanT5XL ViT-L (3.4B) & 68.87 & 64.03\dec{4.84} & \second{58.91}\dec{9.96} & 60.96\dec{7.91} & \best{51.20}\dec{17.67} \\
\midrule
\multirow{2}{*}{\makecell[c]{EVA-CLIP\\ViT-G/14}}
& BLIP-2 FlanT5XL (4.1B) & 70.17 & 64.37\dec{5.80} & 62.45\dec{7.72} & \second{59.93}\dec{10.24} & \best{51.77}\dec{18.40} \\
& InstructBLIP FlanT5XL (4.1B) & 72.87 & 66.77\dec{6.10} & 60.63\dec{12.24} & \second{59.13}\dec{13.74} & \best{53.35}\dec{19.52} \\
\bottomrule
\end{tabular}
\end{table*}

\begin{table*}[t]
\centering
\caption{VQA results. Scores are GPT-5-judged answer correctness rates on a 0--100 scale; down arrows denote absolute drops from clean outputs. Adapted baselines are defined and cited in Table~\ref{tab:cls_nat}.}
\label{tab:vqa_results}
\footnotesize
\setlength{\tabcolsep}{3.8pt}
\renewcommand{\arraystretch}{1.08}
\begin{tabular}{C{1.85cm} C{3.95cm} *{4}{C{1.52cm}} G{1.68cm}}
\toprule
\textbf{Image Encoder} & \textbf{Model} & \textbf{Clean} & \textbf{QR Code} & \textbf{HCB} & \textbf{AdvGrid} & \oursheader \\
\midrule
\multirow[c]{4}{*}{\makecell[c]{OpenAI CLIP\\ViT-L/14}}
  & LLaVA-1.5 (7B) & 68.21 & 63.34\dec{4.87} & 58.56\dec{9.65} & \second{57.28}\dec{10.93} & \best{51.34}\dec{16.87} \\
  & LLaVA-1.6 (13B) & 71.38 & 64.23\dec{7.15} & 61.65\dec{9.73} & \second{59.36}\dec{12.02} & \best{55.23}\dec{16.15} \\
  & OpenFlamingo (3B) & 51.56 & 41.85\dec{9.71} & \second{35.74}\dec{15.82} & 37.46\dec{14.10} & \best{32.03}\dec{19.53} \\
  & BLIP-2 FlanT5XL ViT-L (3.4B) & 62.32 & 58.12\dec{4.20} & 57.78\dec{4.54} & \second{55.78}\dec{6.54} & \best{46.78}\dec{15.54} \\
\midrule
\multirow[c]{2}{*}{\makecell[c]{EVA-CLIP\\ViT-G/14}}
  & BLIP-2 FlanT5XL (4.1B) & 65.83 & \second{57.58}\dec{8.25} & 60.56\dec{5.27} & 59.34\dec{6.49} & \best{47.34}\dec{18.49} \\
  & InstructBLIP FlanT5XL (4.1B) & 64.78 & 59.10\dec{5.68} & 57.39\dec{7.39} & \second{55.21}\dec{9.57} & \best{42.23}\dec{22.55} \\
\bottomrule
\end{tabular}
\end{table*}
\runinhead{Captioning and VQA.} We further evaluate whether the same physical patch disrupts open-ended language outputs, not only closed-set classification decisions. The captioning results are reported in Table~\ref{tab:captioning_results}, and the VQA results are reported in Table~\ref{tab:vqa_results}. Captioning measures whether the generated description remains semantically consistent with the image, whereas VQA measures whether the model can still answer a question correctly. The supplementary material provides the fixed judge prompts and score aggregation rules: caption scores are normalized from a 0--30 rubric to a 0--100 scale, while VQA scores are binary judge decisions averaged over evaluated question-image pairs and reported on a 0--100 scale. A strong attack should therefore reduce both scores across different language heads and image encoders.

\begin{figure*}[t]
\centering
\includegraphics[width=0.96\textwidth]{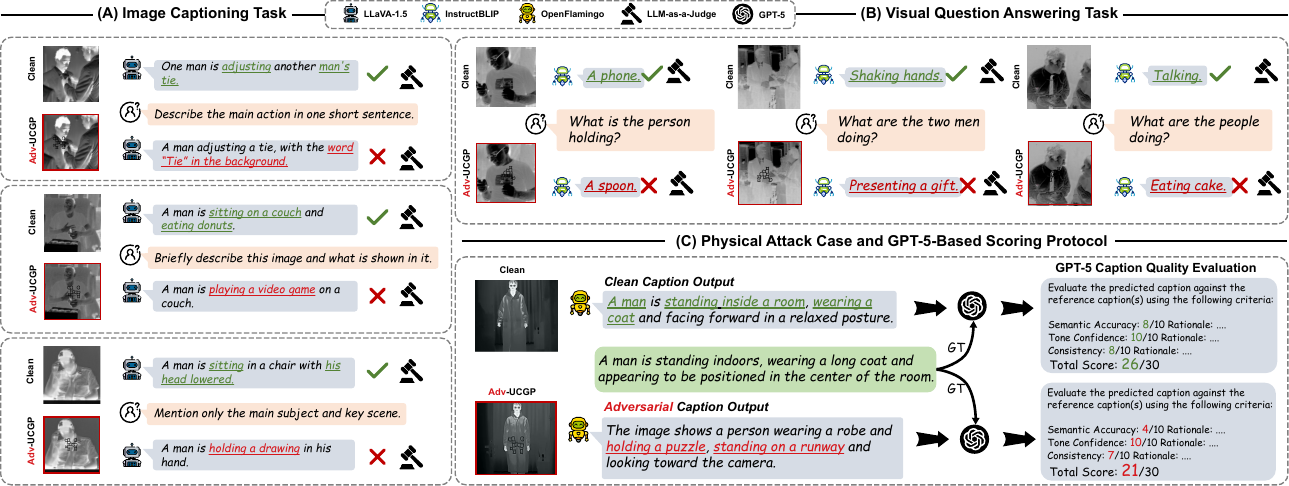}
\caption{Semantic failures of IR-VLMs in captioning, VQA, and physical deployment under the proposed patch. (A) Captioning errors. (B) VQA errors. (C) Physical examples with GPT-5-based scoring. Colored words mark \correct{correct semantics} and \wrong{semantic errors}.}
\label{fig:caption_vqa_examples}
\end{figure*}

Tables~\ref{tab:captioning_results} and~\ref{tab:vqa_results} show that \ourmethod{} gives the largest degradation in all six captioning and VQA settings. Captioning drops range from 15.24 to 19.52 points, while VQA drops range from 15.54 to 22.55 points. The consistent drops across LLaVA, OpenFlamingo, BLIP-2, and InstructBLIP indicate that the patch affects the shared visual evidence used by different language interfaces rather than only one decoder. Fig.~\ref{fig:caption_vqa_examples} provides qualitative examples of these failures: the adversarial outputs remain fluent, but they often change the subject category, action, or scene relation. This supports the quantitative finding that the attack causes semantic degradation rather than only minor wording changes. The agreement between the quantitative drops and output-level errors shows that the patch-induced representation shift propagates consistently to downstream language generation and question answering. This cross-task consistency reduces the likelihood that the degradation is metric- or model-specific.

\subsection{Physical Attack Experiments}
\label{sec:physical-attack-experiments}

\begin{figure*}[!t]
\centering
\includegraphics[width=0.94\textwidth]{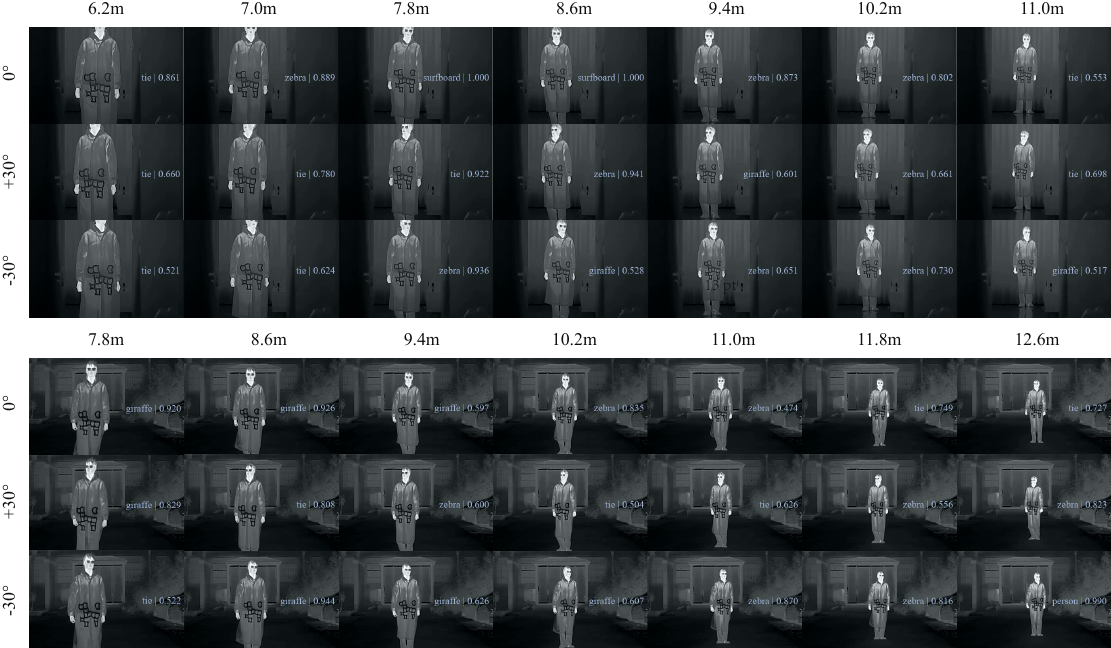}
\caption{Physical attack examples across distances and viewpoints. Indoor examples cover 6.2--11.0 m, while outdoor examples cover the farther 7.8--12.6 m range enabled by the open scene. The patch remains visible in infrared and changes the predicted semantics under both deployments.}
\label{fig:physical_examples}
\end{figure*}

\begin{figure}[!t]
\centering
\includegraphics[width=0.98\columnwidth]{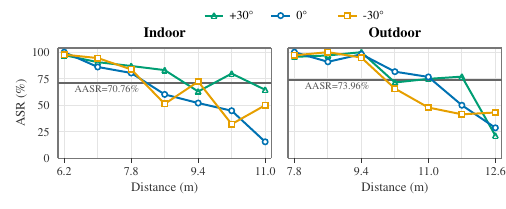}
\caption{ASR trends of the physical attack under indoor and outdoor deployment. The indoor curve uses the 6.2--11.0 m range, and the outdoor curve uses the farther 7.8--12.6 m range shown in Fig.~\ref{fig:physical_examples}. Horizontal reference lines denote the scene-level average ASR, abbreviated AASR.}
\label{fig:physical_asr}
\end{figure}

We emphasize that \ourmethod{} is not designed to hide the wearer or remain invisible to infrared sensors; it is a visible physical patch designed to induce semantic errors in IR-VLMs. To verify real-scene feasibility, we conduct physical deployment experiments by attaching the optimized patch to the target clothing region and capturing infrared videos from multiple distances and viewpoints in indoor and outdoor scenes. The supplementary material provides the submitted video filenames and metadata, and all reported physical ASRs are computed on the 3-fps uniformly sampled frames described in Sec.~\ref{sec:experimental-setup}. As shown in Fig.~\ref{fig:physical_examples}, the indoor examples cover 6.2--11.0 m, while the outdoor examples are intentionally shifted to a farther range of 7.8--12.6 m to stress-test longer-distance deployment in a more open scene. The patch remains visible in infrared and induces semantic deviations across both distance ranges and viewing angles. Fig.~\ref{fig:physical_asr} further quantifies this trend, yielding average ASRs of 70.76\% indoors and 73.96\% outdoors. Although outdoor evaluation uses a farther distance range, its slightly higher average attack success rate (AASR) may be partly explained by the more complex outdoor thermal background: surrounding structures and non-target objects can provide additional visual cues that make the IR-VLM representation less cleanly centered on the person, so the patch-induced semantic shift can more easily move the prediction toward competing categories. At the same time, the farthest outdoor cases still show degradation because long-range capture reduces the effective patch resolution and makes thermal contrast more sensitive to cloth deformation, pose, and background temperature. Overall, the attack remains non-trivial across both indoor and outdoor scenes, showing that the learned curved-grid pattern survives the evaluated digital-to-physical transition rather than relying on a purely digital artifact.

\subsection{Cross-Dataset Transfer}
\label{sec:cross-dataset-transfer}

In real-world deployment, the training and test domains are often mismatched. We therefore optimize the shared patch on one infrared dataset and apply it directly to others without fine-tuning. Table~\ref{tab:cross-dataset-transfer} reports ASR, $\Delta s_{\mathrm{prom}}$, and $M_{\mathrm{adv}}$ across all source--target pairs. The results show non-trivial cross-dataset transfer, with the strongest transfer generally appearing on OpenAI CLIP ViT-L/14 and weaker but still measurable transfer on OpenCLIP ViT-B/16, Meta-CLIP ViT-L/14, and EVA-CLIP ViT-G/14.

\begin{table*}[!t]
\centering
\begingroup
\setlength{\tabcolsep}{2.4pt}
\renewcommand{\arraystretch}{0.98}
\caption{Cross-dataset transfer results. For each surrogate/source pair, ASR, $\Delta s_{\mathrm{prom}}$, and $M_{\mathrm{adv}}$ are reported on the target datasets and their average (AVG). The score-movement descriptors follow the definition in Sec.~\ref{sec:experimental-setup}. \srcdomain{1}--\srcdomain{5} denote Infrared-COCO, LSOTB-TIR, LLVIP, M3FD, and FLIR ADAS v1.3. Gray, boldface diagonal cells indicate the no-transfer setting in which the source and target datasets coincide.}
\label{tab:cross-dataset-transfer}
\scriptsize
\resizebox{\textwidth}{!}{%
\begin{tabular}{@{}cccccccccccccccccccc@{}}
\toprule
\multirow{2}{*}{Surrogate Models} & \multirow{2}{*}{Source} & \multicolumn{3}{c}{Infrared-COCO} & \multicolumn{3}{c}{LSOTB-TIR} & \multicolumn{3}{c}{LLVIP} & \multicolumn{3}{c}{M3FD} & \multicolumn{3}{c}{FLIR v1.3} & \multicolumn{3}{c}{AVG} \\
\cmidrule(lr){3-5}\cmidrule(lr){6-8}\cmidrule(lr){9-11}\cmidrule(lr){12-14}\cmidrule(lr){15-17}\cmidrule(lr){18-20}
 &  & ASR & $\Delta s_{\mathrm{prom}}$ & $M_{\mathrm{adv}}$ & ASR & $\Delta s_{\mathrm{prom}}$ & $M_{\mathrm{adv}}$ & ASR & $\Delta s_{\mathrm{prom}}$ & $M_{\mathrm{adv}}$ & ASR & $\Delta s_{\mathrm{prom}}$ & $M_{\mathrm{adv}}$ & ASR & $\Delta s_{\mathrm{prom}}$ & $M_{\mathrm{adv}}$ & ASR & $\Delta s_{\mathrm{prom}}$ & $M_{\mathrm{adv}}$ \\
\midrule
\multirow{5}{*}{\shortstack{OpenAI CLIP\\ViT-L/14}} & \srcdomain{1} & \diagcell{94.33} & \diagcell{30.80} & \diagcell{34.86} & 98.53 & 31.29 & 21.81 & 93.36 & 23.99 & 21.25 & 100.00 & 27.78 & 27.59 & 100.00 & 56.64 & 52.31 & 97.24 & 34.10 & 31.56 \\
 & \srcdomain{2} & 81.05 & 45.25 & 40.31 & \diagcell{100.00} & \diagcell{25.69} & \diagcell{20.18} & 96.46 & 22.82 & 25.11 & 100.00 & 36.51 & 31.35 & 100.00 & 60.34 & 50.86 & 95.50 & 38.12 & 33.56 \\
 & \srcdomain{3} & 81.70 & 37.30 & 38.96 & 99.37 & 28.81 & 26.18 & \diagcell{96.46} & \diagcell{28.41} & \diagcell{23.06} & 99.49 & 35.50 & 32.24 & 100.00 & 57.62 & 56.94 & 95.40 & 37.53 & 35.48 \\
 & \srcdomain{4} & 86.27 & 36.09 & 35.74 & 100.00 & 33.29 & 25.80 & 97.35 & 25.63 & 21.10 & \diagcell{100.00} & \diagcell{31.29} & \diagcell{25.92} & 100.00 & 56.50 & 59.78 & 96.72 & 36.56 & 33.67 \\
 & \srcdomain{5} & 81.70 & 34.21 & 36.80 & 100.00 & 25.76 & 23.80 & 95.13 & 29.09 & 26.83 & 100.00 & 37.69 & 28.76 & \diagcell{100.00} & \diagcell{57.87} & \diagcell{52.31} & 95.37 & 36.92 & 33.70 \\
\midrule
\multirow{5}{*}{\shortstack{OpenCLIP\\ViT-B/16}} & \srcdomain{1} & \diagcell{61.67} & \diagcell{34.15} & \diagcell{34.01} & 23.64 & 21.25 & 13.45 & 26.40 & 28.41 & 21.91 & 21.91 & 16.77 & 15.91 & 14.81 & 28.97 & 27.77 & 29.69 & 25.91 & 22.61 \\
 & \srcdomain{2} & 21.47 & 41.23 & 38.98 & \diagcell{18.06} & \diagcell{26.56} & \diagcell{21.62} & 38.40 & 34.42 & 21.96 & 13.45 & 25.27 & 21.31 & 17.39 & 3.40 & 1.14 & 21.75 & 26.18 & 21.00 \\
 & \srcdomain{3} & 22.98 & 28.38 & 27.27 & 21.11 & 25.62 & 18.29 & \diagcell{50.81} & \diagcell{29.69} & \diagcell{19.00} & 8.04 & 22.03 & 20.11 & 22.73 & 16.51 & 34.64 & 25.13 & 24.45 & 23.86 \\
 & \srcdomain{4} & 12.27 & 27.72 & 25.68 & 15.54 & 24.73 & 19.64 & 32.93 & 32.78 & 26.88 & \diagcell{41.67} & \diagcell{19.86} & \diagcell{22.40} & 12.00 & 16.11 & 17.66 & 22.88 & 24.24 & 22.45 \\
 & \srcdomain{5} & 14.19 & 28.85 & 26.14 & 19.73 & 30.13 & 21.84 & 39.09 & 29.36 & 26.78 & 21.05 & 15.02 & 21.06 & \diagcell{35.71} & \diagcell{13.84} & \diagcell{15.16} & 25.95 & 23.44 & 22.20 \\
\midrule
\multirow{5}{*}{\shortstack{Meta-CLIP\\ViT-L/14}} & \srcdomain{1} & \diagcell{34.33} & \diagcell{42.58} & \diagcell{44.76} & 46.58 & 35.45 & 27.93 & 46.09 & 37.32 & 22.60 & 32.60 & 27.17 & 20.56 & 12.12 & 2.31 & 1.89 & 34.34 & 28.97 & 23.55 \\
 & \srcdomain{2} & 70.21 & 44.42 & 36.38 & \diagcell{82.69} & \diagcell{30.31} & \diagcell{18.83} & 84.33 & 48.99 & 35.40 & 18.64 & 37.47 & 24.94 & 30.00 & 11.28 & 6.08 & 57.17 & 34.49 & 24.33 \\
 & \srcdomain{3} & 12.08 & 49.28 & 42.13 & 14.89 & 29.92 & 19.41 & \diagcell{43.87} & \diagcell{44.21} & \diagcell{32.08} & 24.44 & 26.90 & 23.10 & 16.13 & 11.10 & 16.55 & 22.28 & 32.28 & 26.65 \\
 & \srcdomain{4} & 12.93 & 43.25 & 39.50 & 16.00 & 31.28 & 22.65 & 27.30 & 43.27 & 33.73 & \diagcell{32.60} & \diagcell{33.14} & \diagcell{25.55} & 21.70 & 69.65 & 42.97 & 22.11 & 44.12 & 32.88 \\
 & \srcdomain{5} & 22.98 & 40.66 & 34.89 & 19.73 & 28.50 & 20.31 & 26.40 & 48.02 & 37.33 & 32.60 & 31.83 & 26.63 & \diagcell{40.00} & \diagcell{19.94} & \diagcell{11.84} & 28.34 & 33.79 & 26.20 \\
\midrule
\multirow{5}{*}{\shortstack{EVA-CLIP\\ViT-G/14}} & \srcdomain{1} & \diagcell{48.33} & \diagcell{36.85} & \diagcell{30.48} & 46.84 & 39.77 & 26.85 & 51.42 & 33.80 & 27.08 & 16.77 & 51.21 & 51.21 & 11.76 & 31.18 & 24.32 & 35.02 & 38.56 & 31.99 \\
 & \srcdomain{2} & 45.00 & 37.94 & 30.83 & \diagcell{80.45} & \diagcell{33.07} & \diagcell{24.22} & 82.08 & 36.11 & 29.27 & 65.38 & 52.99 & 47.62 & 50.00 & 29.14 & 31.64 & 64.58 & 37.85 & 32.72 \\
 & \srcdomain{3} & 20.54 & 34.47 & 34.62 & 34.02 & 35.27 & 25.24 & \diagcell{50.69} & \diagcell{35.18} & \diagcell{22.08} & 32.24 & 52.71 & 52.14 & 14.29 & 37.25 & 30.59 & 30.36 & 38.98 & 32.93 \\
 & \srcdomain{4} & 12.82 & 34.91 & 28.82 & 21.54 & 38.83 & 24.99 & 33.21 & 33.97 & 23.70 & \diagcell{26.67} & \diagcell{50.91} & \diagcell{48.75} & 12.50 & 35.32 & 31.26 & 21.35 & 38.79 & 31.50 \\
 & \srcdomain{5} & 14.16 & 33.33 & 34.59 & 25.69 & 42.53 & 28.84 & 32.48 & 38.11 & 32.35 & 56.60 & 49.11 & 49.10 & \diagcell{71.43} & \diagcell{23.34} & \diagcell{30.92} & 40.07 & 37.28 & 35.16 \\
\bottomrule
\end{tabular}%
}
\endgroup
\end{table*}
Table~\ref{tab:cross-dataset-transfer} reveals a clear backbone-dependent transfer pattern. OpenAI CLIP is the most stable surrogate: its average cross-dataset ASR remains between 95.37\% and 97.24\% for all five source datasets, indicating that the learned perturbation survives substantial changes in scene composition, sensor characteristics, and thermal appearance. In contrast, OpenCLIP yields consistently lower averages (21.75--29.69\%), whereas Meta-CLIP (22.11--57.17\%) and EVA-CLIP (21.35--64.58\%) vary more strongly with the source domain. This variation is consistent with differences in backbone pretraining and in the geometry of their infrared-domain representations, rather than with a uniform transfer effect across encoders. The score descriptors provide complementary evidence for those images whose predictions change successfully: larger $\Delta s_{\mathrm{prom}}$ and $M_{\mathrm{adv}}$ indicate a more pronounced movement toward the promoted decision. Taken together, the results establish strong cross-dataset transfer for OpenAI CLIP and measurable, but domain-sensitive, transfer for the remaining backbones. This variation therefore motivates evaluation across backbones. Reporting these dimensions separately clarifies whether a transfer gain reflects robust patch behavior or backbone-specific alignment.

\FloatBarrier

\subsection{Cross-Model Transfer}
\label{sec:cross-model-transfer}
Cross-model transfer evaluates whether the learned patch exploits idiosyncrasies of one visual encoder or induces a representation shift that can survive across different CLIP-family backbones. This experiment is complementary to cross-dataset transfer: the source dataset may vary as in Table~\ref{tab:cross-dataset-transfer}, but the key variable here is the victim model used at evaluation time. For each surrogate model and source dataset, we optimize one UCGP patch only on the surrogate backbone and then apply the same patch directly to OpenAI CLIP ViT-L/14, OpenCLIP ViT-B/16, Meta-CLIP ViT-L/14, and EVA-CLIP ViT-G/14 without victim-specific fine-tuning or re-optimization. Thus, the diagonal entries measure same-backbone effectiveness, whereas the off-diagonal entries measure true cross-model transfer.

The transfer asymmetry is also informative. OpenAI CLIP and EVA-CLIP sources tend to transfer more strongly than Meta-CLIP sources, whose average victim ASR ranges from 12.74\% to 28.01\%. This suggests that differences in pretraining data, architecture scale, and infrared adaptation behavior lead to different local geometry around the person representation. Overall, the results support partial cross-model transfer while also showing that IR-VLM attack evaluation should report heterogeneous victim backbones rather than relying on a single surrogate model. ASR provides the primary evidence of cross-model transfer, while the two score metrics further characterize the successful prediction changes.

\begin{table*}[!t]
\centering
\begingroup
\setlength{\tabcolsep}{2.4pt}
\renewcommand{\arraystretch}{0.98}
\caption{Cross-model transfer results. For each surrogate/source pair, ASR, $\Delta s_{\mathrm{prom}}$, and $M_{\mathrm{adv}}$ are reported on the victim models and their average (AVG). The score-movement descriptors follow the definition in Sec.~\ref{sec:experimental-setup}. Source labels follow Table~\ref{tab:cross-dataset-transfer}. Gray, boldface diagonal cells indicate the no-transfer setting in which the surrogate and victim models coincide.}
\label{tab:cross-model-transfer}
\scriptsize
\resizebox{\textwidth}{!}{%
\begin{tabular}{@{}ccccccccccccccccc@{}}
\toprule
\multirow{2}{*}{Surrogate Models} & \multirow{2}{*}{Source} & \multicolumn{3}{c}{OpenAI CLIP ViT-L/14} & \multicolumn{3}{c}{OpenCLIP ViT-B/16} & \multicolumn{3}{c}{Meta-CLIP ViT-L/14} & \multicolumn{3}{c}{EVA-CLIP ViT-G/14} & \multicolumn{3}{c}{AVG} \\
\cmidrule(lr){3-5}\cmidrule(lr){6-8}\cmidrule(lr){9-11}\cmidrule(lr){12-14}\cmidrule(lr){15-17}
 &  & ASR & $\Delta s_{\mathrm{prom}}$ & $M_{\mathrm{adv}}$ & ASR & $\Delta s_{\mathrm{prom}}$ & $M_{\mathrm{adv}}$ & ASR & $\Delta s_{\mathrm{prom}}$ & $M_{\mathrm{adv}}$ & ASR & $\Delta s_{\mathrm{prom}}$ & $M_{\mathrm{adv}}$ & ASR & $\Delta s_{\mathrm{prom}}$ & $M_{\mathrm{adv}}$ \\
\midrule
\multirow{5}{*}{\shortstack{OpenAI CLIP\\ViT-L/14}} & \srcdomain{1} & \diagcell{94.33} & \diagcell{30.80} & \diagcell{34.86} & 14.53 & 31.50 & 28.98 & 51.15 & 132.89 & 68.24 & 13.95 & 81.73 & 48.40 & 43.49 & 69.23 & 45.12 \\
 & \srcdomain{2} & \diagcell{100.00} & \diagcell{25.69} & \diagcell{20.18} & 7.76 & 60.10 & 36.95 & 9.22 & 32.66 & 34.02 & 7.98 & 28.83 & 25.09 & 31.24 & 36.82 & 29.06 \\
 & \srcdomain{3} & \diagcell{96.46} & \diagcell{28.41} & \diagcell{23.06} & 18.58 & 43.79 & 35.18 & 38.50 & 98.92 & 72.14 & 22.58 & 59.24 & 48.46 & 44.03 & 57.59 & 44.71 \\
 & \srcdomain{4} & \diagcell{100.00} & \diagcell{31.29} & \diagcell{25.92} & 14.72 & 49.52 & 45.46 & 45.19 & 99.75 & 85.44 & 19.81 & 43.48 & 43.18 & 44.93 & 56.01 & 50.00 \\
 & \srcdomain{5} & \diagcell{100.00} & \diagcell{57.87} & \diagcell{52.31} & 33.34 & 29.76 & 28.44 & 25.94 & 39.78 & 34.61 & 44.44 & 77.83 & 42.36 & 50.93 & 51.31 & 39.43 \\
\midrule
\multirow{5}{*}{\shortstack{OpenCLIP\\ViT-B/16}} & \srcdomain{1} & 36.75 & 72.16 & 37.15 & \diagcell{61.67} & \diagcell{34.15} & \diagcell{34.01} & 11.38 & 30.79 & 19.51 & 15.56 & 22.74 & 15.37 & 31.34 & 39.96 & 26.51 \\
 & \srcdomain{2} & 14.67 & 20.21 & 18.75 & \diagcell{18.06} & \diagcell{26.56} & \diagcell{21.62} & 14.36 & 11.81 & 13.93 & 12.43 & 16.26 & 18.06 & 14.88 & 18.71 & 18.09 \\
 & \srcdomain{3} & 31.22 & 28.17 & 27.37 & \diagcell{50.81} & \diagcell{29.69} & \diagcell{19.00} & 40.14 & 50.68 & 43.05 & 33.07 & 35.14 & 32.50 & 38.81 & 35.92 & 30.48 \\
 & \srcdomain{4} & 4.19 & 15.01 & 16.69 & \diagcell{41.67} & \diagcell{19.86} & \diagcell{22.40} & 6.75 & 15.52 & 11.10 & 4.11 & 12.97 & 7.81 & 14.18 & 15.84 & 14.50 \\
 & \srcdomain{5} & 37.89 & 8.07 & 8.20 & \diagcell{35.71} & \diagcell{13.84} & \diagcell{15.16} & 16.42 & 10.56 & 14.57 & 16.42 & 19.45 & 19.11 & 26.61 & 12.98 & 14.26 \\
\midrule
\multirow{5}{*}{\shortstack{Meta-CLIP\\ViT-L/14}} & \srcdomain{1} & 9.43 & 65.66 & 40.91 & 9.94 & 56.39 & 43.81 & \diagcell{34.33} & \diagcell{42.58} & \diagcell{44.76} & 12.34 & 43.21 & 32.88 & 16.51 & 51.96 & 40.59 \\
 & \srcdomain{2} & 10.63 & 35.25 & 29.48 & 9.17 & 21.82 & 14.75 & \diagcell{82.69} & \diagcell{30.31} & \diagcell{18.83} & 9.55 & 23.50 & 22.54 & 28.01 & 27.72 & 21.40 \\
 & \srcdomain{3} & 9.71 & 38.06 & 24.13 & 11.63 & 52.12 & 38.47 & \diagcell{43.87} & \diagcell{44.21} & \diagcell{32.08} & 10.47 & 49.29 & 35.44 & 18.92 & 45.92 & 32.53 \\
 & \srcdomain{4} & 11.99 & 49.10 & 35.32 & 5.68 & 33.81 & 25.87 & \diagcell{32.60} & \diagcell{33.14} & \diagcell{25.55} & 5.21 & 34.43 & 32.10 & 13.87 & 37.62 & 29.71 \\
 & \srcdomain{5} & 3.26 & 3.34 & 6.75 & 6.07 & 3.04 & 12.39 & \diagcell{40.00} & \diagcell{19.94} & \diagcell{11.84} & 1.63 & 1.60 & 9.98 & 12.74 & 6.98 & 10.24 \\
\midrule
\multirow{5}{*}{\shortstack{EVA-CLIP\\ViT-G/14}} & \srcdomain{1} & 46.60 & 90.56 & 53.92 & 5.74 & 56.33 & 44.84 & 19.53 & 48.90 & 33.68 & \diagcell{48.33} & \diagcell{36.85} & \diagcell{30.48} & 30.05 & 58.16 & 40.73 \\
 & \srcdomain{2} & 7.61 & 23.82 & 22.72 & 8.14 & 17.78 & 13.64 & 8.20 & 8.97 & 9.38 & \diagcell{80.45} & \diagcell{33.07} & \diagcell{24.22} & 26.10 & 20.91 & 17.49 \\
 & \srcdomain{3} & 14.94 & 22.31 & 16.98 & 16.16 & 36.17 & 32.63 & 22.09 & 61.90 & 51.63 & \diagcell{50.69} & \diagcell{35.18} & \diagcell{22.08} & 25.97 & 38.89 & 30.83 \\
 & \srcdomain{4} & 7.45 & 12.59 & 10.19 & 9.22 & 5.37 & 3.54 & 9.86 & 15.01 & 8.44 & \diagcell{26.67} & \diagcell{50.91} & \diagcell{48.75} & 13.30 & 20.97 & 17.73 \\
 & \srcdomain{5} & 7.49 & 98.37 & 103.42 & 21.14 & 7.20 & 10.05 & 51.34 & 37.85 & 46.09 & \diagcell{71.43} & \diagcell{23.34} & \diagcell{30.92} & 37.85 & 41.69 & 47.62 \\
\bottomrule
\end{tabular}%
}
\endgroup
\end{table*}

Table~\ref{tab:cross-model-transfer} reports ASR, $\Delta s_{\mathrm{prom}}$, and $M_{\mathrm{adv}}$ for every surrogate--victim pairing. The same-backbone diagonal is generally strongest, reaching 94.33--100.00\% for OpenAI CLIP sources and 26.67--80.45\% for EVA-CLIP sources, which confirms that UCGP can reliably attack the model used during optimization. Off-diagonal transfer is weaker but still observable. For example, OpenAI CLIP surrogates obtain average victim ASR between 31.24\% and 50.93\%. These results demonstrate measurable but heterogeneous cross-model transfer, with substantially lower ASR for several heterogeneous surrogate--victim pairs.

\subsection{Cross-Category Extension}
\label{sec:cross-category-extension}

We evaluate cross-category extension to test whether UCGP is a person-specific attack or a category-agnostic framework that can be re-instantiated for other infrared objects. This setting differs from cross-dataset and cross-model transfer: the dataset and model family are kept fixed, but the target semantic category is changed. Therefore, the experiment asks whether the same design principles--a clean-category reference structure, curved-grid patch parameterization, representation-level objective, and derivative-free optimization--remain useful when the clean manifold corresponds to a different object shape and thermal appearance.

For each target category, we follow the same protocol used in the main person experiment. We first select clean Infrared-COCO samples that the evaluated classifier stably recognizes as the target category, then extract their visual features to rebuild the category-specific clean reference structure. A single shared UCGP patch is optimized for that category, with the same patch scale rule, loss weights, EOT/TPS augmentation strategy, query budget, and evaluation metrics as in the main setting. We do not introduce category-specific redesigns such as hand-tuned shapes, different objectives, or different placement rules. After optimization, adversarial samples are evaluated by classification ASR, promoted-class score change $\Delta s_{\text{prom}}$, and adversarial margin $M_{\text{adv}}$, so the table reports both final label flips and conditional score movement.

\begin{table}[tp]
\centering
\caption{Cross-category extension results on Infrared-COCO. Each category is evaluated on $N=300$ samples. ASR is reported in percent; $\Delta s_{\text{prom}}$ and $M_{\text{adv}}$ are the score-movement descriptors defined in Sec.~\ref{sec:experimental-setup}.}
\label{tab:cross-category-extension}
\footnotesize
\setlength{\tabcolsep}{5.0pt}
\renewcommand{\arraystretch}{1.06}
\begin{tabular}{L{1.65cm} C{1.30cm} C{1.42cm} C{1.42cm}}
\toprule
\textbf{Category} & \textbf{ASR (\%)} & $\boldsymbol{\Delta s_{\textbf{prom}}}$ & $\boldsymbol{M_{\textbf{adv}}}$ \\
\midrule
Person & 94.33 & 98.75 & 69.32 \\
Dog & \best{98.67} & \best{106.25} & \best{76.43} \\
Bicycle & 56.00 & 95.12 & 44.62 \\
Car & 61.00 & 92.32 & 50.87 \\
\bottomrule
\end{tabular}
\end{table}

Table~\ref{tab:cross-category-extension} shows that UCGP remains effective beyond the main person setting, but the attack strength depends on category geometry and thermal structure. Dog reaches 98.67\% ASR, slightly higher than the person result of 94.33\%, suggesting that compact animal-like infrared shapes can be strongly disturbed by a localized curved-grid pattern. Car and bicycle obtain lower ASRs of 61.00\% and 56.00\%, respectively. A likely reason is that these categories are more rigid, elongated, or spatially extended, so a fixed-size local patch covers a smaller fraction of the discriminative object evidence and leaves more clean category cues outside the patched region. The accompanying conditional score descriptors summarize the promoted-class behavior for those samples that cross the decision boundary. These results support the claim that UCGP is not restricted to person samples, while also clarifying an important boundary: category-specific shape, thermal contrast, and effective patch coverage modulate the achievable attack strength.

\subsection{Ablation and Design Analysis}
\label{sec:ablation-studies}
\begin{figure}[!t]
\centering
\includegraphics[width=0.92\columnwidth]{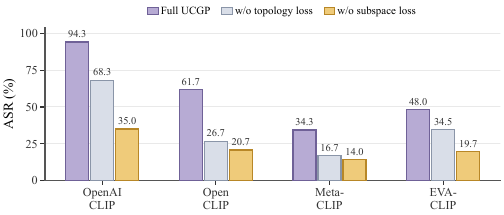}
\caption{Loss-term ablation results for selected objective components. The bars compare full \ourmethod{} with variants that remove the topology loss or the subspace loss, providing an extended objective-level view that complements the module-wise comparison in Table~\ref{tab:module_ablation}.}
\label{fig:module_ablation_bars}
\end{figure}
\begin{figure}[!t]
\centering
\includegraphics[width=0.92\columnwidth]{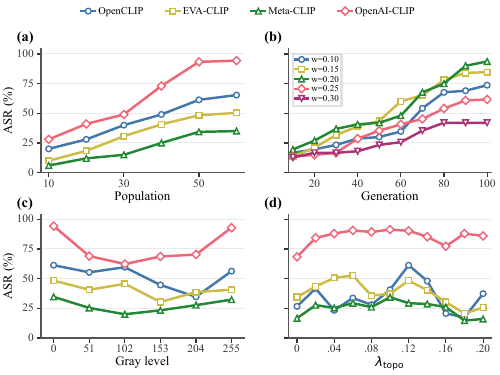}
\caption{Parameter sensitivity of key design choices. (a) Population size. (b) Convergence under grid-line widths. (c) Patch color. (d) Graph-KL weight $\lambda_{\text{topo}}$.}
\label{fig:ablation_curves}
\vspace{-10pt}
\end{figure}
\begin{figure}[tp]
\centering
\includegraphics[width=0.92\columnwidth]{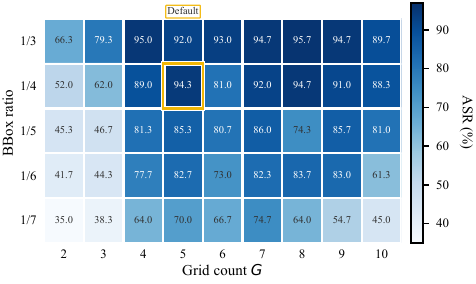}
\caption{Heatmap comparison of the patch-area ratio (patch area divided by the target bounding-box area) and grid count. Each cell reports ASR in percent; the gold outline marks the default configuration used in the main experiments reported here.}
\label{fig:bbox_grid_heatmap}
\vspace{-10pt}
\end{figure}

We conduct ablation and design analysis to identify which components of UCGP are responsible for the attack effect. All variants use the same Infrared-COCO person evaluation split, patch scale, placement rule, query budget, and classification ASR metric as the main classification experiment. The only changed factor is the component named in the corresponding row or curve. This controlled protocol separates four design questions: whether the curved-grid parameterization is necessary, whether curved connectivity is better than straight or random structures, whether MetaDE improves the derivative-free search, and whether the subspace and topology losses each contribute to the final semantic shift. Table~\ref{tab:module_ablation} reports module-level replacements, Fig.~\ref{fig:module_ablation_bars} isolates objective-loss removals, and Fig.~\ref{fig:ablation_curves} together with Fig.~\ref{fig:bbox_grid_heatmap} examines hyperparameter sensitivity. Together, these controlled comparisons attribute performance changes to structural parameterization, objective design, optimization behavior, or hyperparameter selection under a shared protocol.

\begin{table}[!t]
\centering
\caption{Module-wise ablation results. ASR is in percent, where higher indicates a stronger attack. Each variant changes only the named component; Pixel-wise patch directly optimizes pixels within the attack region instead of using the curved-grid parameterization.}
\label{tab:module_ablation}
\footnotesize
\setlength{\tabcolsep}{4.0pt}
\renewcommand{\arraystretch}{1.06}
\begin{tabular}{@{}lcccc@{}}
\toprule
\textbf{Variant} &
\makecell{\textbf{OpenAI}\\\textbf{CLIP}} &
\makecell{\textbf{Open-}\\\textbf{CLIP}} &
\makecell{\textbf{Meta-}\\\textbf{CLIP}} &
\makecell{\textbf{EVA-}\\\textbf{CLIP}} \\
\midrule
\best{Full \ourmethod{}} & \best{94.33} & \best{61.67} & \best{34.33} & \best{48.00} \\
Pixel-wise patch & 30.33 & 21.67 & 13.33 & 12.33 \\
\ourmethod{} w/ straight segments & 7.00 & 5.00 & 3.67 & 4.33 \\
\ourmethod{} w/ random topology & 5.33 & 3.67 & 2.00 & 4.00 \\
w/o MetaDE, std. DE & 76.00 & 54.67 & 30.33 & 46.33 \\
\bottomrule
\end{tabular}
\end{table}

\begin{figure}[H]
\centering
\includegraphics[width=0.92\columnwidth]{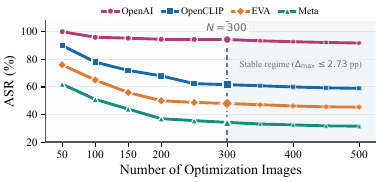}
\caption{Effect of the number of optimization images on classification ASR across four CLIP backbones. After 300 images, the ASR curves approximately converge across backbones.}
\label{fig:image_count_ablation}
\end{figure}

The module and loss ablations show that the structured patch representation and the representation-level objective are both important. In Fig.~\ref{fig:module_ablation_bars}, removing the topology loss lowers ASR from 94.33\% to 68.33\% on OpenAI CLIP ViT-L/14, and removing the subspace loss further lowers it to 35.00\%; the same ordering holds on the other backbones, showing that topology preservation helps but the subspace term provides the strongest representation-displacement signal. In Table~\ref{tab:module_ablation}, replacing the curved-grid parameterization with direct pixel-wise patch optimization reduces ASR from 94.33\% to 30.33\% on OpenAI CLIP ViT-L/14 and produces similar drops on the other backbones. Replacing curved edges with straight segments or random topologies further reduces ASR to near-failure levels, indicating that a deployable low-frequency pattern alone is insufficient: the connected curved structure is needed to generate a stable semantic perturbation under the same physical constraints. Replacing MetaDE with standard DE also lowers ASR across all backbones, but the drop is smaller than the structural ablations, suggesting that the optimizer improves search quality while the curved-grid representation contributes the dominant structural advantage.

The parameter analyses further check whether the default configuration is robust or merely cherry-picked. Fig.~\ref{fig:ablation_curves} varies population size, compares convergence under different grid-line widths, patch color, and the topology weight $\lambda_{\text{topo}}$ while keeping the remaining settings fixed. The curves show that a medium population size, black patch color, moderate topology weight, and medium line width give the most stable ASR, which is consistent with the physical intuition that the patch must be visible enough to change infrared evidence but not so large or dense that it becomes hard to attach stably. Fig.~\ref{fig:bbox_grid_heatmap} reaches the same conclusion from patch scale and grid density: too small a region or too sparse a grid weakens the attack signal, whereas an overly large or dense pattern introduces deployment difficulty and less stable gains. Fig.~\ref{fig:image_count_ablation} shows that, across all four backbones, the ASR curves approximately converge after 300 optimization images; we therefore use 300 images in the main experiments. Together, these ablations support UCGP as a coupled design rather than a collection of independent tricks: the curved-grid structure constrains the searchable patch family, the representation losses define the semantic direction of the perturbation, and MetaDE searches this constrained space effectively.

\section{Defense Evaluation}
\label{sec:defense-evaluation}

To evaluate system-level security, we test adversarial training~\cite{goodfellow2015fgsm} and digital watermarking~\cite{hayes2018visible}, abbreviated AT and DW. DW is an image-repair defense: DW-1 is blind to the patch location, while DW-2 uses the patch-location prior for guided restoration. For AT, we construct the hardened classification set from training samples whose retained class label contains person, generate adversarially augmented samples using randomly transformed UCGP patch renderings from the same EOT/TPS transformation family, and fine-tune only the LoRA parameters while keeping the VLM backbone frozen. We then test the original patch on the hardened model as AT-1 and adaptively re-optimize the patch against it as AT-2. Table~\ref{tab:defense} summarizes the results.

\begin{table}[!t]
\centering
\caption{Defense evaluation. ASR is reported in percent; lower ASR indicates a stronger defense. Drop denotes the absolute ASR reduction relative to No Defense; -- marks the reference row.}
\label{tab:defense}
\footnotesize
\setlength{\tabcolsep}{4.0pt}
\renewcommand{\arraystretch}{1.10}
\begin{tabular}{L{1.55cm} L{2.35cm} C{1.05cm} C{1.05cm}}
\toprule[0.75pt]
\textbf{Setting} & \textbf{Defense access} & \makecell[c]{\textbf{ASR}\\\textbf{(\%)}} & \makecell[c]{\textbf{Drop}\\\textbf{(\%)}} \\
\midrule[0.45pt]
\mbox{No Defense} & None & 94.33 & -- \\
DW-1 & Blind repair & 45.33 & 49.00 \\
DW-2 & Patch-location prior & \second{24.67} & \second{69.66} \\
AT-1 & Original patch & \best{15.33} & \best{79.00} \\
AT-2 & Adaptive patch & 78.00 & 16.33 \\
\bottomrule[0.75pt]
\end{tabular}
\end{table}    

The defense results show that current defenses can mitigate the attack, but none provides a complete defense. DW-1 reduces ASR from 94.33\% to 45.33\%, and DW-2 further reduces it to 24.67\% by using the patch-location prior, indicating that guided local restoration is more effective than blind repair but still incomplete. AT-1 gives the lowest ASR of 15.33\% because the hardened model has directly seen the original universal patch distribution. However, AT-2 rebounds to 78.00\% after adaptive patch re-optimization, showing that robustness against a fixed patch does not eliminate the underlying semantic vulnerability. Defense evaluation should therefore include both known-patch and adaptively re-optimized settings, and robust IR-VLM defense may need to combine image-level repair with representation-level training.
\section{Discussion}
\label{sec:discussion}

The central implication of this study is that the exposed vulnerability is representational rather than only classificatory. In infrared imagery, local thermal contrast and contour-like structures can carry strong semantic evidence because texture and color cues are weaker than in visible images. A deployable curved-grid patch can therefore create a locally plausible but misleading structure that shifts the shared visual representation; classification, captioning, and VQA heads then inherit the error in different forms. The transfer results support this view but also bound it: attack strength varies with backbone and dataset geometry, so IR-VLM audits should report source--target matrices and score-movement metrics rather than rely only on a single aggregate ASR.

The design analyses further show that UCGP is not merely adding visible clutter to infrared images. The curved-grid structure in UCGP restricts the search to connected low-frequency curves, while the straight-segment, random-topology, and MetaDE ablations show that curved connectivity and search quality both matter. The main gain therefore comes from coupling a deployable infrared patch prior with a representation-level objective. The remaining limitations define the boundary of the evidence: UCGP assumes a representation-visible audit interface, and EOT/TPS modeling does not fully cover material-dependent thermal response, ambient temperature shifts, sensor nonlinearity, occlusion, or camera post-processing. Future evaluations should include jointly optimized placement, longer videos, multiple thermal cameras, different materials, larger category sets with confidence intervals, and adaptive defenses that combine image-level repair with representation-level robustness training.

\section{Conclusion}
\label{sec:conclusion}

We present a systematic study of universal physical adversarial patch attacks on infrared vision-language models and introduce \ourmethod{}, a universal physical patch framework for this setting. By combining curved-grid parameterization, representation-level objectives, EOT capture sampling, and TPS patch warping, \ourmethod{} learns a single deployable patch that degrades classification, captioning, and VQA, with measurable transfer, cross-category extension, and controlled real-scene effectiveness. These results establish a concrete evaluation setting for physical semantic attacks on IR-VLMs and highlight the need for stronger defenses against cross-modal semantic manipulation in practice.

\section*{Acknowledgments}
This work was supported by the Introduction Program for Young Doctors (`Tianchi Talents') of Xinjiang Uygur Autonomous Region.

\bibliographystyle{IEEEtran}
\bibliography{UCGP_TPAMI_Submission}

\end{document}